\documentclass[lettersize,journal]{IEEEtran}
\usepackage{amsmath,amsfonts}
\usepackage{algorithmic}
\usepackage{algorithm}
\usepackage{array}
\usepackage[caption=false,font=scriptsize,labelfont=sf,textfont=sf]{subfig}
\usepackage{textcomp}
\usepackage{stfloats}
\usepackage{url}
\usepackage{verbatim}
\usepackage{graphicx}
\usepackage{cite}
\usepackage{booktabs}
\usepackage{xcolor}
\definecolor{eccvblue}{rgb}{0.12,0.49,0.85}
\usepackage[pagebackref,breaklinks,colorlinks,citecolor=eccvblue]{hyperref}
\usepackage{multirow}
\usepackage{diagbox}
\usepackage{tikz-imagelabels}

\newcommand\fft{\mathcal{F}}
\newcommand\ifft{\mathcal{F}^{-1}}
\begin{document}

\title{A Hybrid Transformer-Mamba Network for Single Image Deraining} 

\author{Shangquan Sun, Wenqi Ren, Juxiang Zhou, Jianhou Gan, Rui Wang, Xiaochun Cao,~\IEEEmembership{Senior Member,~IEEE}

\thanks{This paper was produced by the IEEE Publication Technology Group. They are in Piscataway, NJ.}
\thanks{Manuscript received April 19, 2021; revised August 16, 2021.}

\thanks{Shangquan Sun and Rui Wang are with the State Key Laboratory of Information Security, Institute of Information Engineering, Chinese Academy of Sciences, Beijing 100093, China. Shangquan Sun is also with the School of Cyber Security, University of Chinese Academy of Sciences, Beijing 100049, China.} 
\thanks{Wenqi Ren and Xiaochun Cao are with the School of Cyber Science and Technology, Shenzhen Campus of Sun Yat-sen University, Shenzhen 518107, China. (E-mail: \href{mailto:renwq3@mail.sysu.edu.cn}{renwq3@mail.sysu.edu.cn}}
\thanks{Juxiang Zhou and Jianhou Gan are with the Key Laboratory of Education Information for Nationalities, Yunnan Normal University, Ministry of Education, Kunming
650031, China.}
}

\markboth{Journal of \LaTeX\ Class Files,~Vol.~14, No.~8, August~2021}%
{Shell \MakeLowercase{\textit{et al.}}: A Sample Article Using IEEEtran.cls for IEEE Journals}


\maketitle

\begin{abstract}
Existing deraining Transformers employ self-attention mechanisms with fixed-range windows or along channel dimensions, limiting the exploitation of non-local receptive fields. 
In response to this issue, we introduce a novel dual-branch hybrid Transformer-Mamba network, denoted as TransMamba, aimed at effectively capturing long-range rain-related dependencies. 
Based on the prior of distinct spectral-domain features of rain degradation and background, we design a spectral-banded Transformer blocks on the first branch.
Self-attention is executed within the combination of the spectral-domain channel dimension to improve the  ability of modeling long-range dependencies.
%
To enhance frequency-specific information,
we present a spectral enhanced feed-forward module that aggregates features in the spectral domain. 
%
In the second branch, Mamba layers are equipped with cascaded bidirectional state space model modules to additionally capture the modeling of both local and global information.
At each stage of both the encoder and decoder, we perform channel-wise concatenation of dual-branch features and achieve feature fusion through channel reduction, enabling more effective integration of the multi-scale information from the Transformer and Mamba branches.
To better reconstruct innate signal-level relations within clean images, we also develop a spectral coherence loss.
Extensive experiments on diverse datasets and real-world images demonstrate the superiority of our method compared against the state-of-the-art approaches. 
We have released the codes and pre-trained models on \href{https://github.com/sunshangquan/TransMamba}{Github}.
\end{abstract}

\begin{IEEEkeywords}
Single image deraining, Rain streak removal, Image restoration, Spectral domain, Transformer, State space model, Hybrid model
\end{IEEEkeywords}    
\begin{figure}[t]
\scriptsize
  \centering
  \subfloat[Input]{
  \begin{minipage}{0.324\linewidth}
    \centering
    \includegraphics[width=1\linewidth]{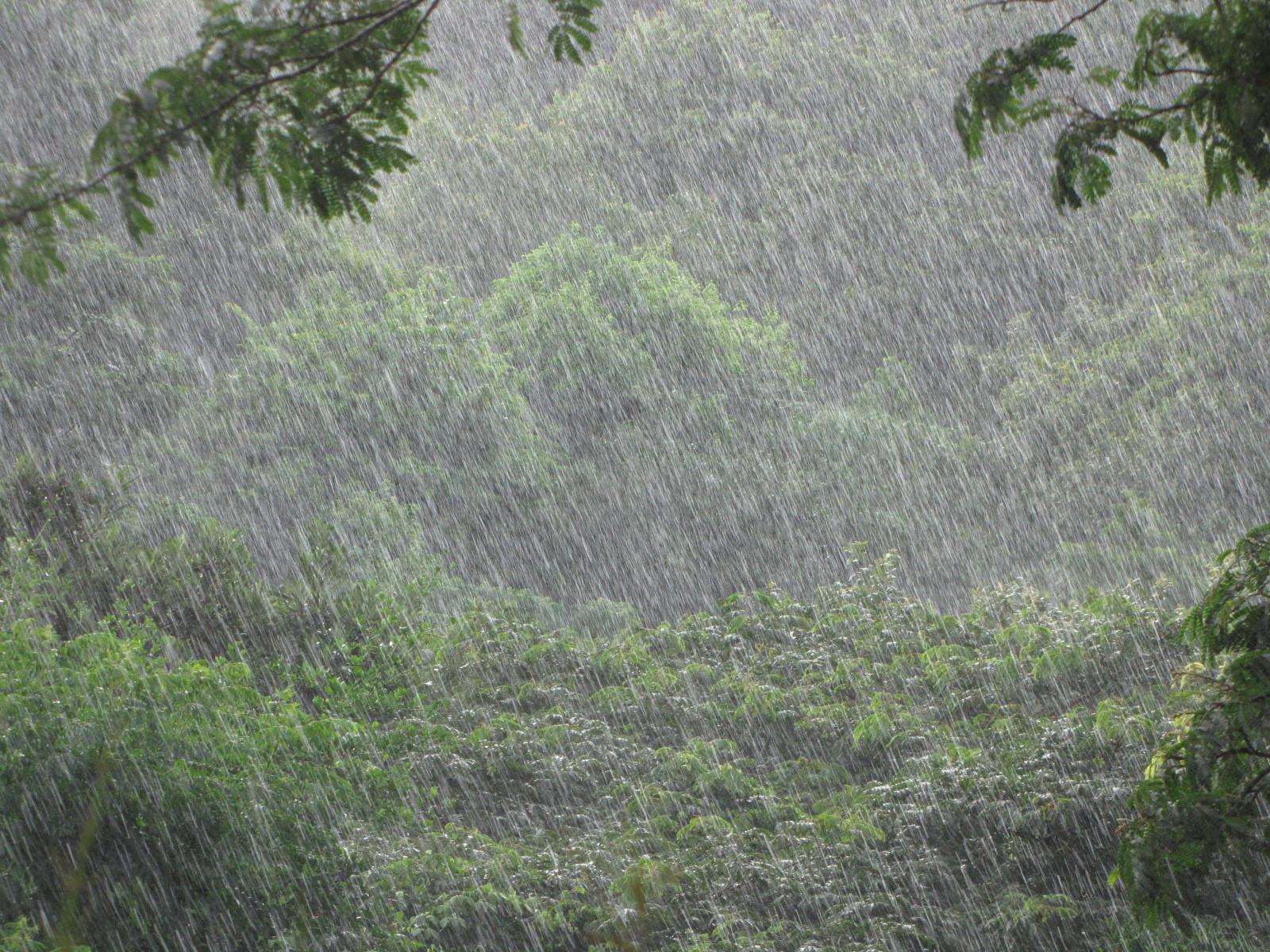}
    \end{minipage}
  }
  \hspace{-2.2mm}
  \subfloat[DualGCN\cite{Fu2021RainSR}]{
  \begin{minipage}{0.324\linewidth}
    \centering
    \includegraphics[width=1\linewidth]{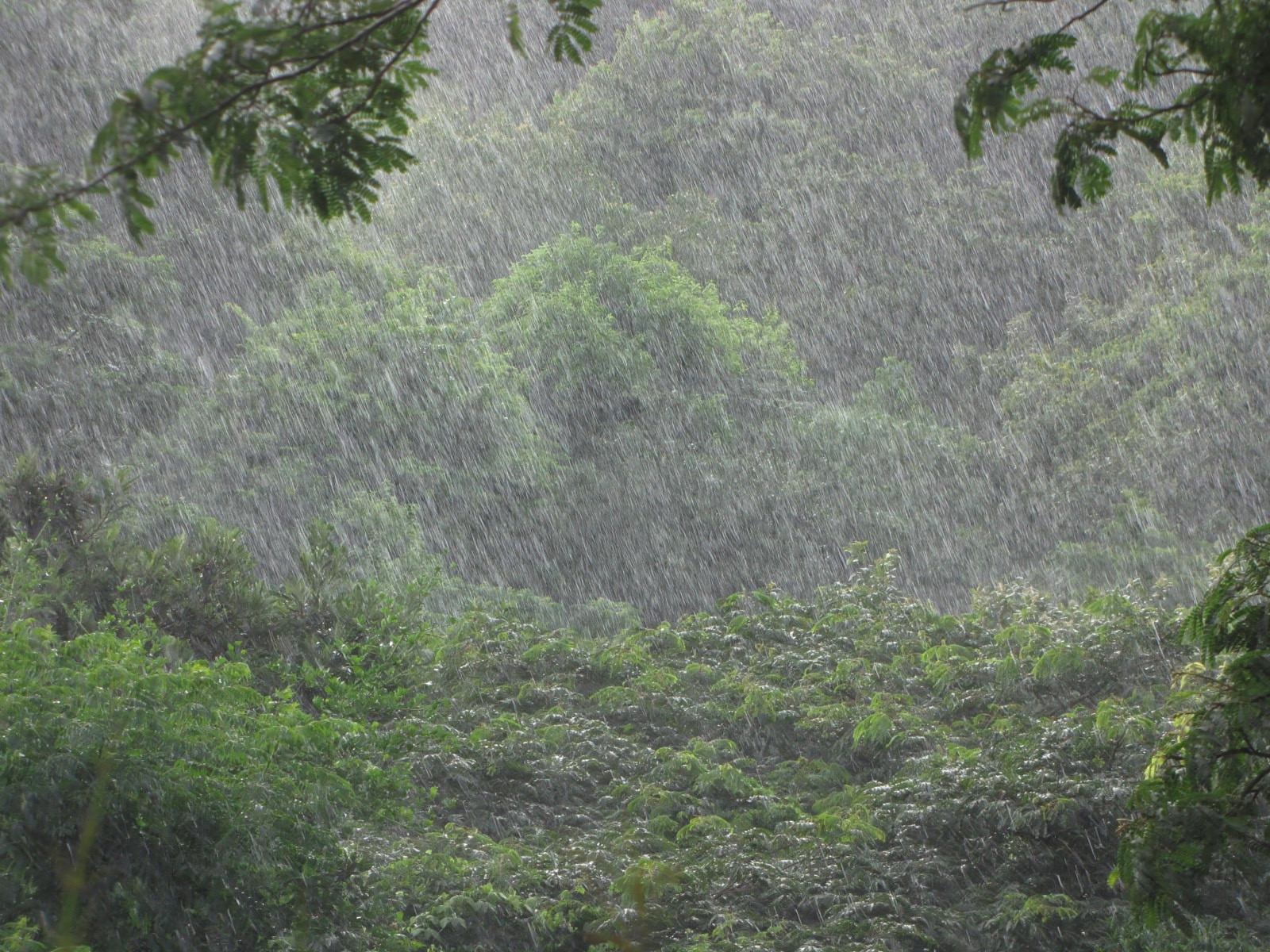}
    \end{minipage}
  }
  \hspace{-2.2mm}
  \subfloat[SPDNet\cite{yi2021Structure}]{
  \begin{minipage}{0.324\linewidth}
    \centering
    \includegraphics[width=1\linewidth]{figs/real/restormer_132.jpg}
    \end{minipage}
  }
  \hspace{-2.2mm}
  \subfloat[Restormer\cite{zamir2022restormer}]{
  \begin{minipage}{0.324\linewidth}
    \centering
    \includegraphics[width=1\linewidth]{figs/real/restormer_132.jpg}
    \end{minipage}
  }
  \hspace{-2.2mm}
  \subfloat[IDT~\cite{xiao2022image}]{
  \begin{minipage}{0.324\linewidth}
    \centering
    \includegraphics[width=1\linewidth]{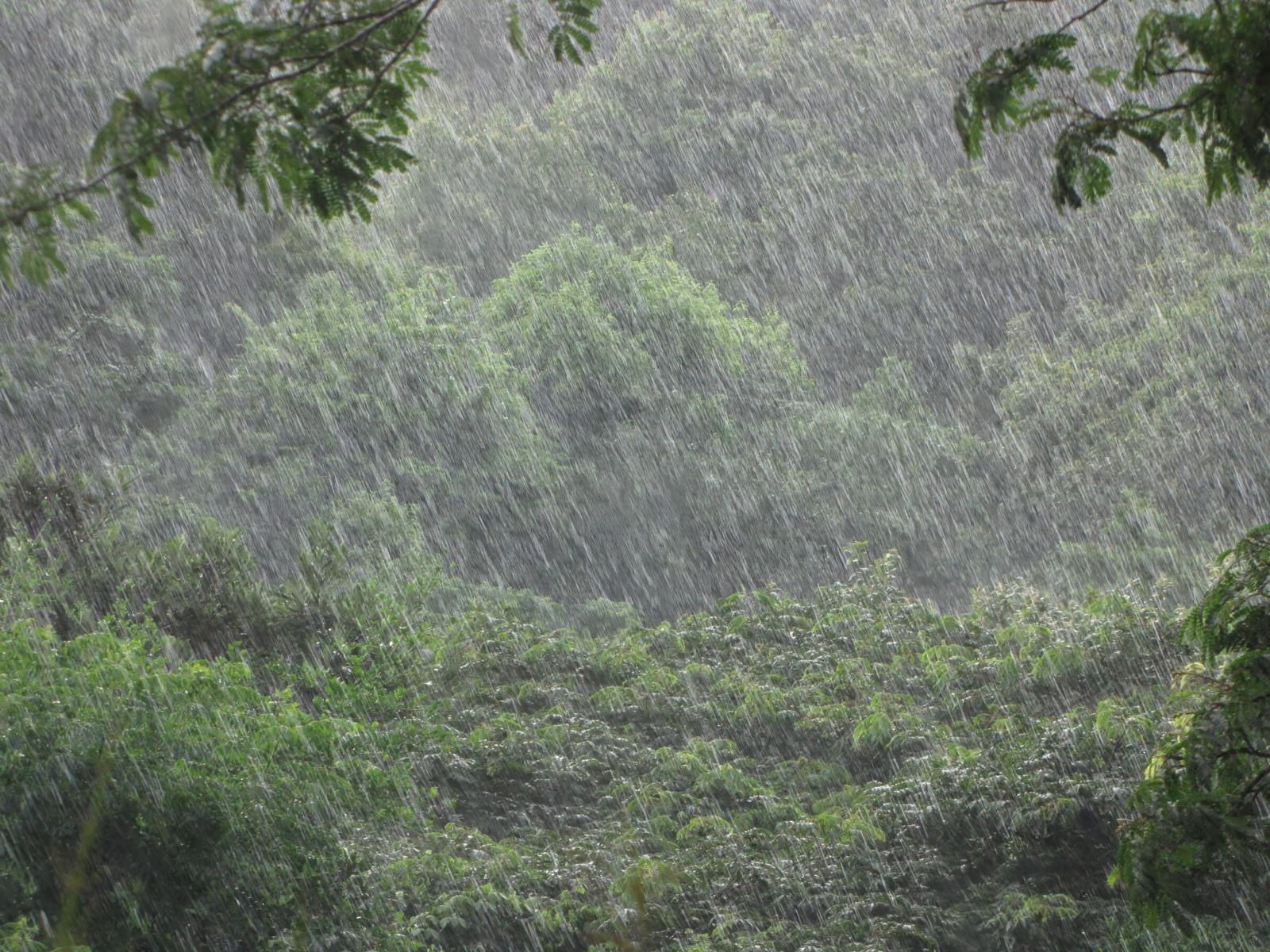}
    \end{minipage}
  }
  \hspace{-2.2mm}
  \subfloat[DRSformer~\cite{chen2023learning}]{
  \begin{minipage}{0.324\linewidth}
    \centering
    \includegraphics[width=1\linewidth]{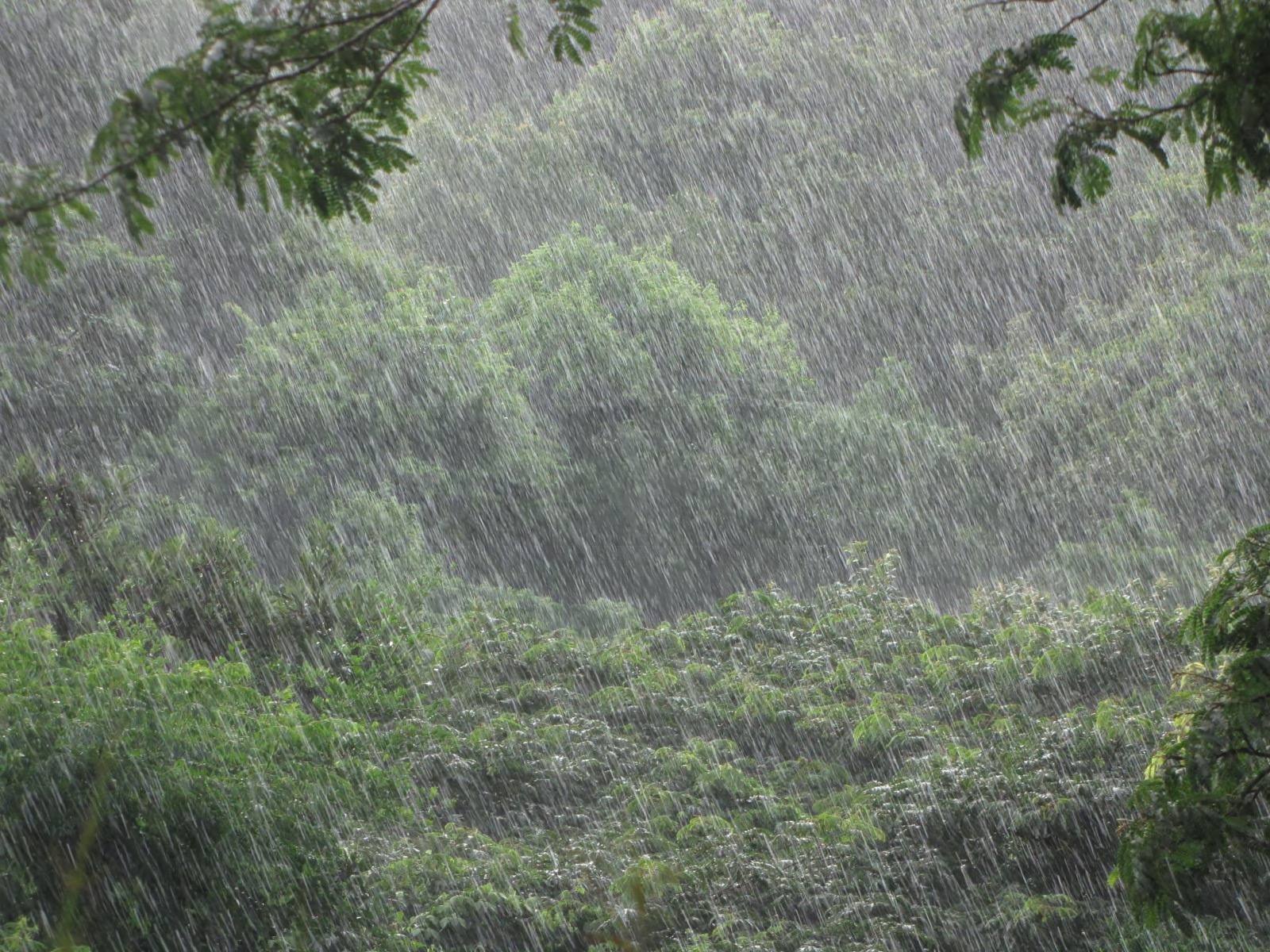}
    \end{minipage}
  }
  \hspace{-2.2mm}
  \subfloat[UDR-S$^2$Former-\cite{chen2023sparse}]{
  \begin{minipage}{0.324\linewidth}
    \centering
    \includegraphics[width=1\linewidth]{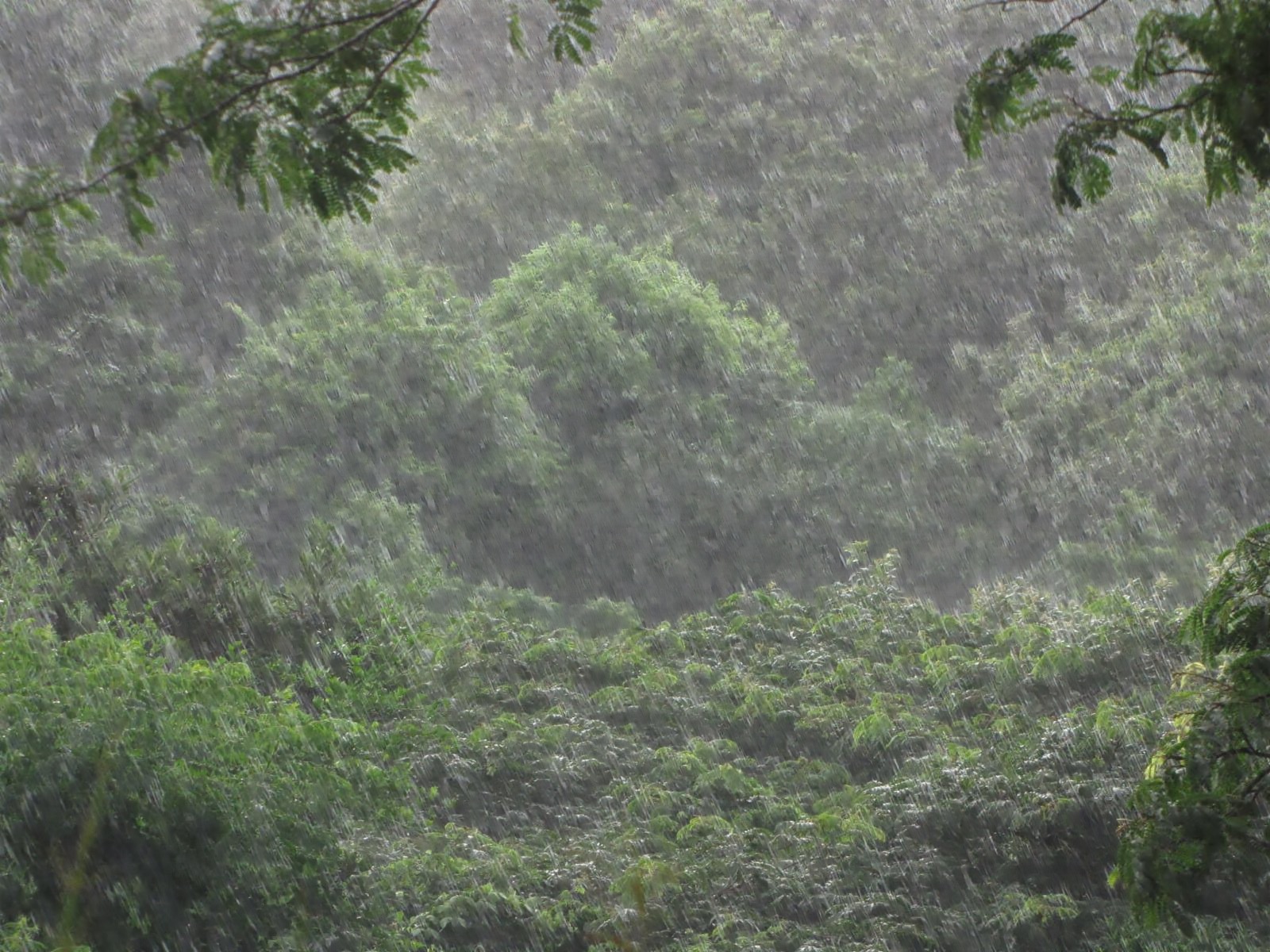}
    \end{minipage}
  }
  \hspace{-2.2mm}
  \subfloat[NeRD-\cite{NeRD-Rain}]{
  \begin{minipage}{0.324\linewidth}
    \centering
    \includegraphics[width=1\linewidth]{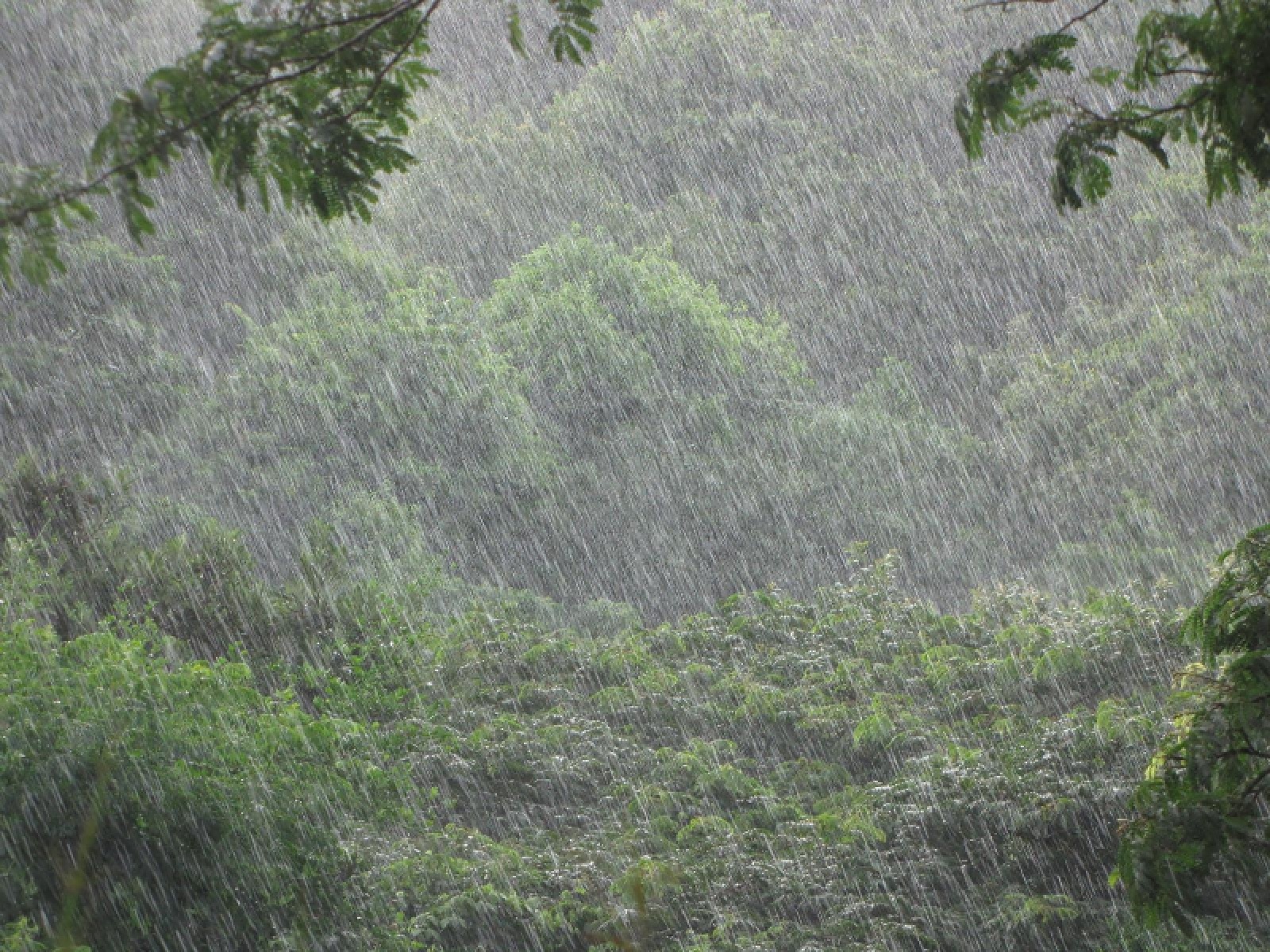}
    \end{minipage}
  }
  \hspace{-2.2mm}
  \subfloat[Ours]{
  \begin{minipage}{0.324\linewidth}
    \centering
    \includegraphics[width=1\linewidth]{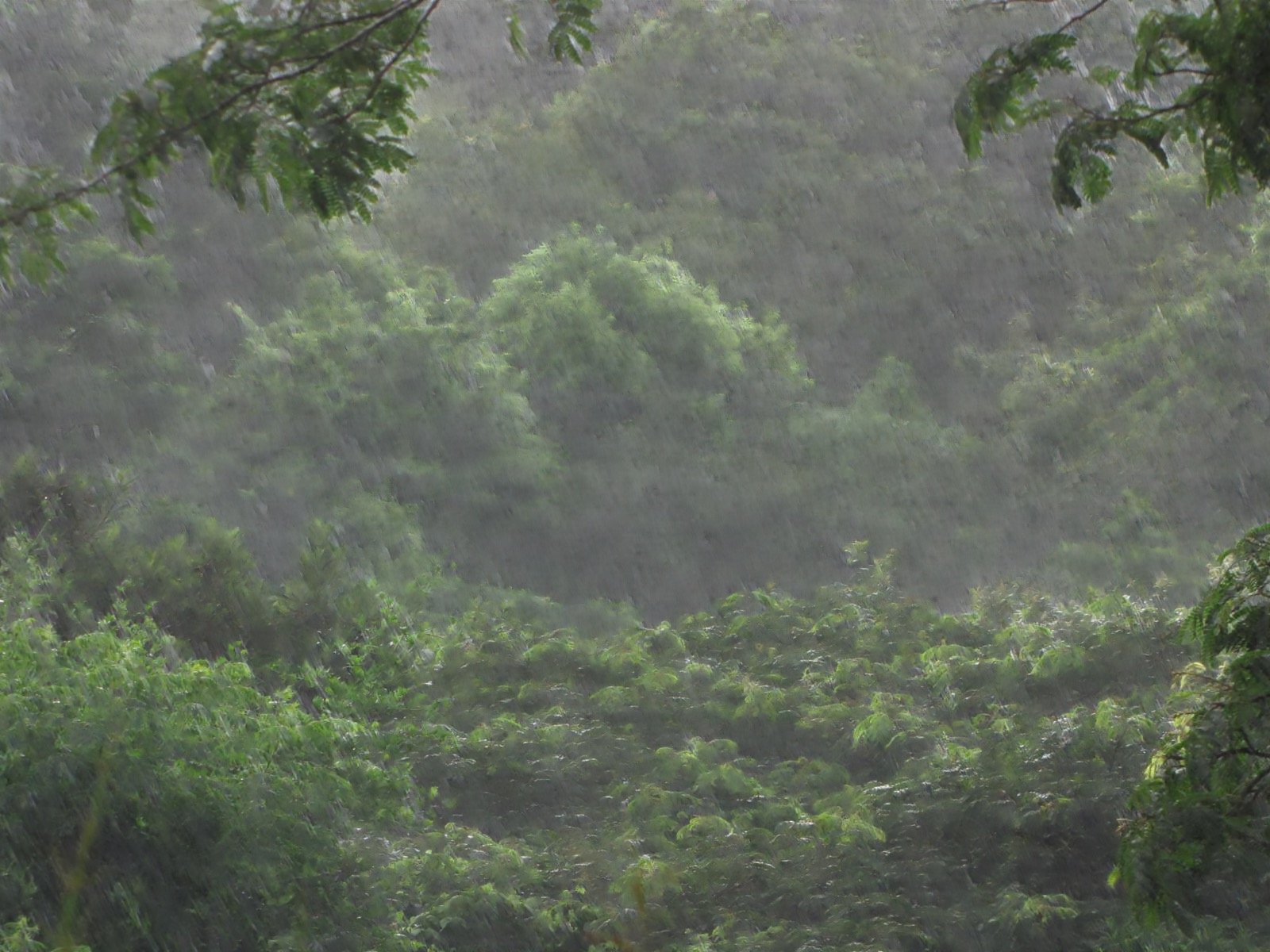}
    \end{minipage}
  }
  \caption{Visual comparisons among the state-of-the-art Transformer deraining methods and ours on real-world rainy Internet-Data~\cite{wang2019spatial}. All model weights for real-world deraining are trained on SPA-Data~\cite{wang2019spatial}. The samples from (b) to (h) are DualGCN~\cite{Fu2021RainSR}, SPDNet~\cite{yi2021Structure}, Restormer~\cite{zamir2022restormer}, IDT~\cite{xiao2022image}, DRSformer~\cite{chen2023learning}, UDR-S$^2$Former~\cite{chen2023sparse}, and NeRD~\cite{NeRD-Rain}. Our method produces the most visually pleasing result on the real-world rainy image.}
  \label{fig:cover2}
\end{figure}
\section{Introduction}
\label{sec:intro}

\looseness-1
Rain, as a common adverse weather, severely degrades outdoor vision and thus rain streak removal is a vital low-level vision task to recover rainy images into clean ones. 
Early works leverage physical priors to design their models such as sparse coding~\cite{kang2011automatic}, low-rank model~\cite{chen2013generalized}, Gaussian mixture model~\cite{li2016rain}, etc. 
However, these models heavily rely on the manual tuning of hyper-parameters and thus cannot remove rains of complex appearances and various scales.

\looseness-1
Recently, convolutional neural networks (CNNs) have outperformed traditional models in modeling complex patterns. 
Various CNN architectures have been developed for single image deraining~\cite{yang2017deep,li2018recurrent,yang2019joint,du2020conditional,jiang2020multi,yasarla2019uncertainty,yasarla2020confidence,yasarla2020syn2real,wei2019semi,huang2021memory,jiang2020multi,ren2019progressive,zamir2021multi}. 
Nonetheless, CNNs fail to learn long-range dependencies due to the limited-range receptive field of convolutional layers. 
To overcome this issue, Transformers~\cite{chen2021pre,zamir2022restormer,chen2023sparse,xiao2022image,chen2023learning,chen2023hybrid,jiang2022magic,sun2024restoring} have been employed to learn non-local representations based on the self-attention mechanism for rain streak removal. 
However, the quadratic computational complexity of the self-attention mechanism compromises the choices of token selection.
The existing deraining Transformers thus require self-attention either restricted within each channel dimension or fixed-size blocks, limiting their ability to fully leverage long-range dependencies. 
The setting also neglects the innate coherence between rain streaks dynamically located but closely correlated in appearance. 

\looseness-1
To address the challenge of efficiently removing rain streaks from images, we propose a dual-branch hybrid Transformer-Mamba (TransMamba) network. Our model consists of two branches: a Transformer branch and a Mamba branch, each serving distinct yet complementary roles in the deraining process.

Rain streaks present unique challenges due to their frequency characteristics in the spectral domain. We observe that each token in the spectral domain represents a sinusoidal component of the original 2D image, offering a globally decomposed representation of specific frequency bands. The lower-frequency components primarily encode the rain streaks, which typically appear as repeated, texture-free areas. In contrast, high-frequency components represent background regions with richer textures.
To leverage this observation, we first transform images into the spectral domain using the Fast Fourier Transform (FFT). This transformation allows us to perform self-attention in the spectral domain, capturing global dependencies more effectively based on frequency decomposition. By isolating the tokens into distinct frequency bands, we introduce spectral banding self-attention (SBSA), which selectively reallocates attention based on the band’s significance. The low-frequency bands containing rain streaks receive reduced attention, while high-frequency bands that encode background textures receive enhanced attention. This selective attention enables effective rain streak removal by attenuating low-frequency rain signals and preserving high-frequency image details.

To further refine feature extraction in the spectral domain, we introduce a spectral enhanced feed-forward (SEFF) module. This module capitalizes on the property that element-wise multiplication in the frequency space is analogous to convolution in the spatial domain. Within our dual-branch architecture, we apply pixel-wise weights and biases to spectral-domain features, enhancing frequency-specific information. The SiLU activation in one branch acts as a gating mechanism, controlling the output of the other branch to achieve optimal spectral enhancement.

While the Transformer branch focuses on capturing global information, our Mamba branch is designed to enhance sequence coherence. Comprising bi-directional state space model (SSM) modules, this branch ensures that linear dependencies between image sequences are preserved. We leverage the coherence measure from signal processing, which quantifies the linear relationship between two signals, to introduce a spectral coherence loss function. This loss function ensures that the reconstructed image maintains the linear relationships inherent in the clean image, resulting in a more accurate and consistent derained output.

In summary, our contributions are in three folds:
\begin{itemize}
    \item We propose a dual-branch hybrid Transformer-Mamba network.
    Its first Transformer branch enables a comprehensive information modeling, conducting attention across spectral-domain tokens into bands of different frequencies. The second Mamba branch is equipped with bi-directional SSM modules to enhance sequence coherence.   
    \item To better extract frequency-specific features, we develop a spectral enhanced feed-forward module. By introducing a spectral coherence loss function, we ensure the reconstruction of the signal-level linear relationship. 
    \item By conducting extensive experiments on various benchmarks and real-world images, we show that our method obtains more favourable deraining performance than the state-of-the-art methods. 
\end{itemize}




\begin{figure}[t]
  \centering
   \includegraphics[width=1\linewidth]{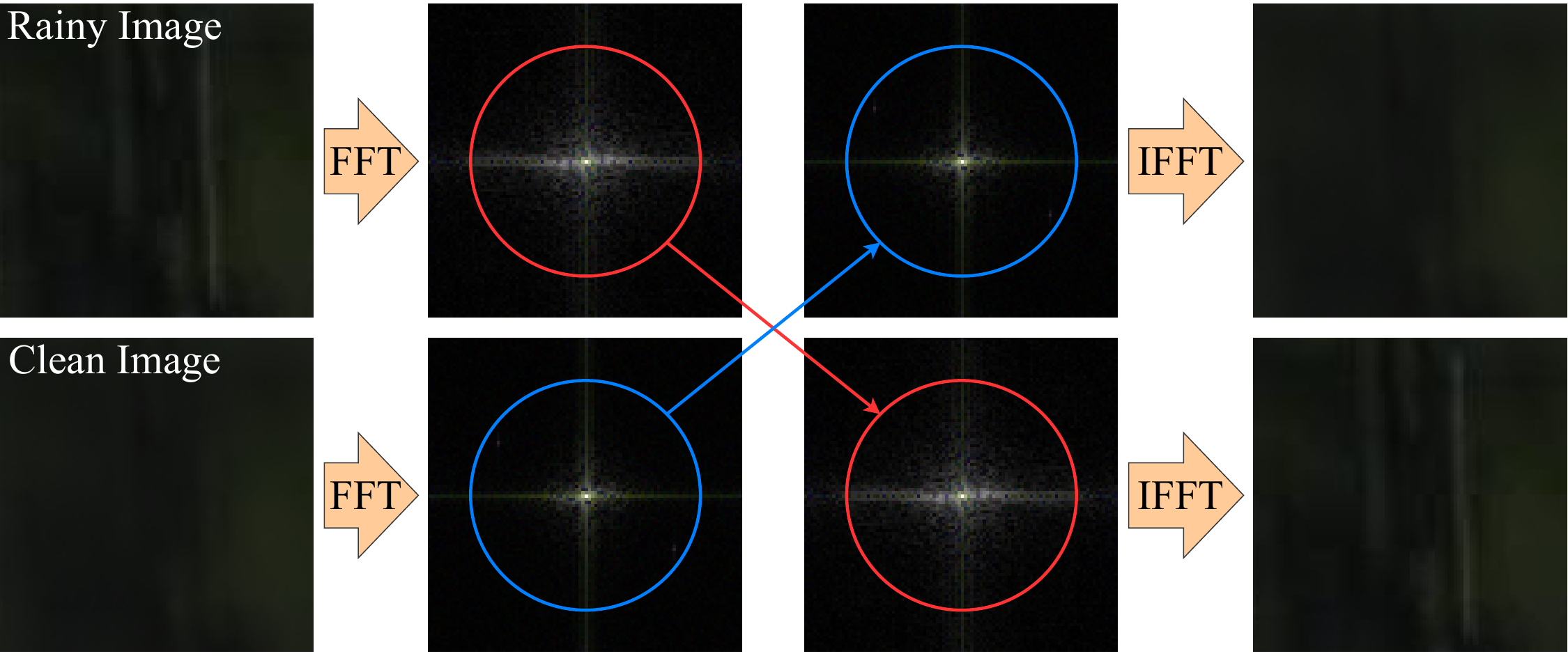}
   \caption{Demonstration that spectral bands of different frequencies separately encode background and rain streaks. 
   The process of replacing the low-frequency band of the rainy signal with that of the clean signal, results in the easy removal of rain streaks. 
   Inspired by the phenomenon, we propose allocating various attention across bands, taking advantage of the distinct information encoding in different frequency bands.}
   \label{fig:intro}
\end{figure}
\section{Related Work}
\label{sec:related}
\noindent\textbf{Single Image Deraining.}
Early methods have tried tackling single image deraining by employing various physical priors of rain streaks~\cite{kang2011automatic,gu2017joint,li2016rain,luo2015removing,zhang2017convolutional}.
But they generally require empirically tuning handcrafted hyper-parameters and cannot adapt to complex rainy scenarios.

In recent years, deep learning methods such as CNNs~\cite{he2016deep,sun2022rethinking} have outperformed traditional methods in many vision tasks including single image deraining~\cite{yang2020single,li2021comprehensive,wang2020single,zhang2023data,chen2023towards,li2019single,quan2021removing}. 
Many networks designs are proposed, such as encoder-decoder~\cite{li2018non}, 
detail network~\cite{fu2017clearing,fu2017removing}, 
recurrent network~\cite{yang2017deep,li2018recurrent,yang2019joint}, 
spiking network~\cite{song2024learning}, 
neural representation learning~\cite{NeRD-Rain}, 
generative adversarial network~\cite{zhang2019image}, 
multi-scale structure~\cite{jiang2020multi,yasarla2019uncertainty,yasarla2020confidence}, 
variational auto-encoder~\cite{du2020conditional}, 
fractal band learning~\cite{yang2020removing}, 
progressive learning~\cite{jiang2020multi,ren2019progressive,zamir2021multi,sun2023event}, and semi-supervised transfer learning~\cite{yasarla2020syn2real,wei2019semi,huang2021memory}. 
Other methods combine deep learning with traditional priors to achieve better deraining effect~\cite{li2019heavy,wang2020model}.
To enable the training of CNN, many datasets have been collected, including those with synthetic rain streaks~\cite{yang2017deep,fu2017removing,zhang2018density} and real-world rain~\cite{wang2019spatial,guo2023sky,ba2022not}. 
Despite surpassing traditional approaches, CNN-based methods have a restricted receptive field and thus are difficult to effectively capture long-range dependencies.

\noindent\textbf{Vision Transformer.} 
More recently, 
Transformers~\cite{vaswani2017attention,devlin2018bert} have been adopted into computer vision and show better performance than CNNs~\cite{dosovitskiy2020image,liu2021swin,touvron2021training,zhu2020deformable,yuan2021tokens,bian2022learnable,lai2024residual}. 
To enable flexible shape of input and scalable inference, other works combine convolution and Transformer~\cite{wu2021cvt,yuan2021incorporating,zhao2024towards}. 
%
For single image deraining, Jiang \textit{et al}.~\cite{jiang2022magic} propose a deraining network unifying CNN and Transformer. 
Xiao \textit{et al}.~\cite{xiao2022image} design a pure Transformer with window-based and spatial self-attention to effectively remove rain streaks. 
To better model rain-induced degradation relationship, sparse Transformers~\cite{chen2023learning,chen2023sparse} have also been proposed. 
However, the existing Transformer-based deraining methods are confined to self-attention along each channel dimension or fixed-range windows, disregarding that rain streaks are diversely located but similar in terms of appearance and thus closely correlated in the spectral domain. 
To this end, we propose a spectral-domain banding Transformer that pays adaptive attention to background features encoded in high frequency components and rain-degraded pixels embedded in low frequency components, and obtain superior deraining performance.

\noindent\textbf{Transformer in the Spectral Domain.}
\looseness-1
Several works explore to combine Transformer architecture and frequency-domain representation to address magnetic resonance super resolution~\cite{fang2022cross}, music information retrieval~\cite{hung2022modeling}, fault diagnosis~\cite{ding2022novel}, video super resolution~\cite{qiu2022learning,qiu2023learning} and deblurring~\cite{kong2023efficient}. 
Some works in hyperspectral image analysis have the name of spectral Transformer but are not related to transforming spatial RGB images into spectral or frequency domain~\cite{li2023spatial,wang2022hyper,he2022dster,song2022bs2t}. 
Fang \textit{et al}.~\cite{fang2022cross} compute the gradient map of input images to embed high-frequency prior before feeding them into Transformer.
Ding at al.~\cite{ding2022novel} apply one-dimensional frequency transformation along time dimension to better encode long-range sequence.
Another work~\cite{qiu2022learning} employs Discrete Cosine Transform within self-attention module, disregarding the complex part of original image signal.
Kong \textit{et al}.~\cite{kong2023efficient} replace the tensor multiplication in the spatial domain within self-attention module with element-wise product among Query, Key and Value patches in frequency domain. 
The existing Transformers in the frequency domain either calculate single dimensional frequency along time dimension for processing sequence data or merely compute patch-level frequency representation, unable to capture global spatial dependencies.
We thus develop an image-level SBSA for global feature extraction and an SEFF to better reweigh extracted information.

\noindent\textbf{Image Restoration in the Spectral Domain.}
Several works have introduced spectral-domain operations to tackle the tasks of image restoration, including super-resolution~\cite{fang2022cross}, video super resolution~\cite{qiu2022learning,qiu2023learning}, deblurring~\cite{kong2023efficient,xint2023freqsel}, and general restoration~\cite{cui2023selective}. 
However, they disregard the the spectral-domain prior of rain streak and thus are not efficiently adaptive to single image deraining.

\noindent\textbf{Vision Mamba.} 
\looseness-1
Most recently, state space models~\cite{smith2022simplified,gu2021mamba,lieber2024jamba,zhu2024visionmamba} like Mamba~\cite{gu2023mamba,hatamizadeh2024mambavision} show their ability in the efficiency of long-range modeling. 
Though several works have explored their ability in vision tasks~\cite{yamashita2024image, guo2024mambair,zhen2024freqmamba,li2024fouriermamba}, some other work points out that attention-based and CNN models are still superior to Mambas in local and short-sequence~\cite{yu2024mambaout}.
To further enhance global modeling capacity,the existing Mamba-based deraining works~\cite{zhen2024freqmamba,li2024fouriermamba,yamashita2024image} all transform images and perform computation in frequency domain by chance. 
In this work, we alternatively experiment on combining CNN, Transformer and Mamba in frequency domain to equip the model with the ability of both local and global modeling.

\section{Method}

%

\subsection{Overall Network Architecture}\label{sec::overall_arch}

\begin{figure*}[t]
  \centering
   \includegraphics[width=1\linewidth]{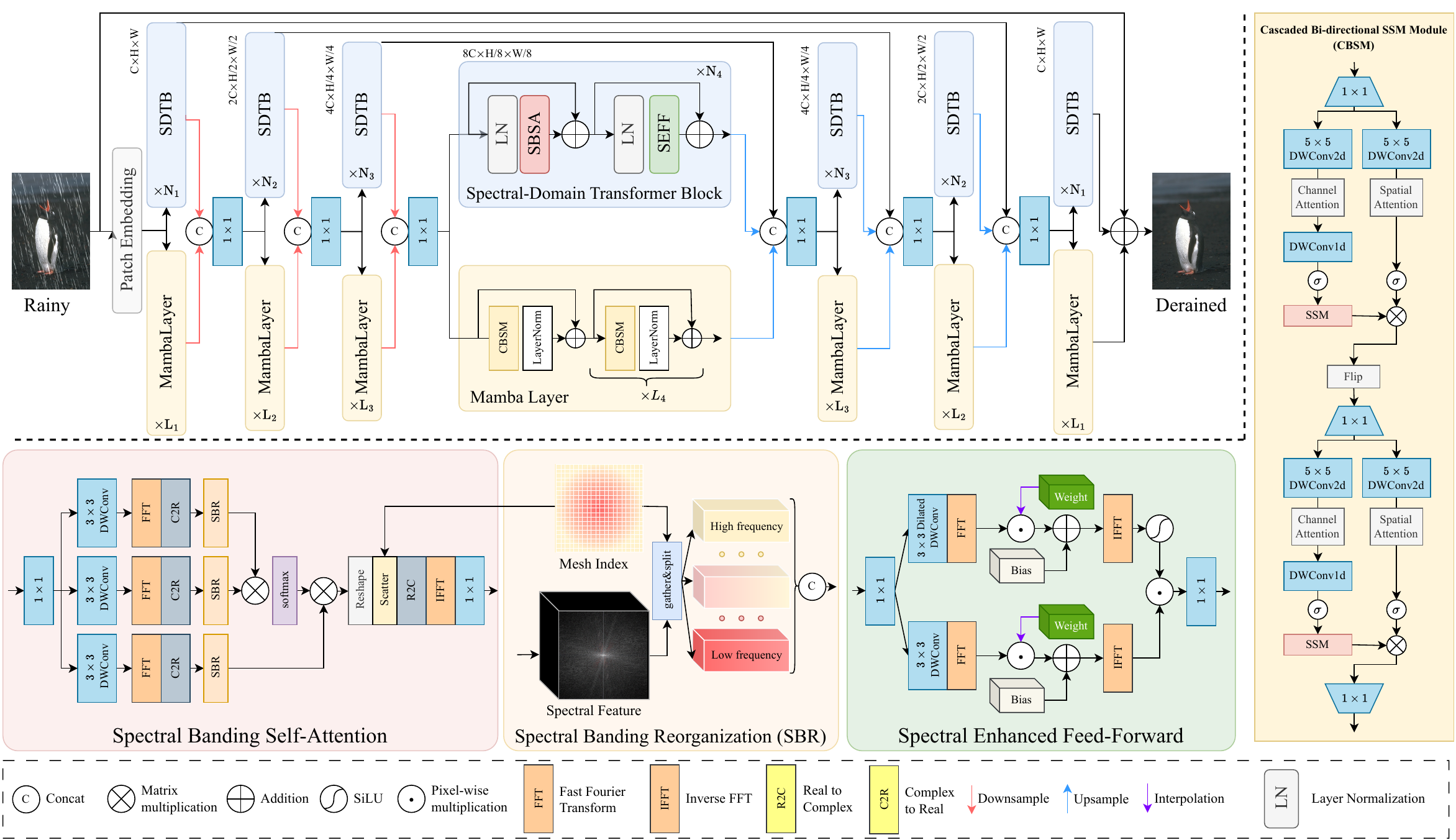}
   \caption{The architecture of our hybrid Transformer-Mamba network (TransMamba) follows a dual-branch structure containing four levels. 
   Each level consists of \(\rm N_i\) spectral-domain Transformer Blocks (SDTBs) and \(\rm L_i\) spatial-domain Mamba layers. 
   Each SDTB is composed of spectral banding self-attention (SBSA) and spectral enhanced feed-forward (SEFF). 
   Within SBSA, we present spectral banding reorganization (SBR) to categorize high and low frequency features.
   Each Mamba layer contains multiple cascaded Bi-directional SSM modules (CBSMs).
   }
   \label{fig:structure}
\end{figure*}

The general network architecture of our proposed TransMamba, illustrated in Figure~\ref{fig:structure}, follows a symmetrical multi-level dual-branch encoder-decoder structure. 
Consider the input as a rainy image denoted as \(R\in \mathbb{R}^{3\times H\times W}\), where \(H\) and \(W\) represent the image's height and width, respectively. 
We employ a \(3\times 3\) convolution to generate the overlapping embedding of image patches.
Within the blocks in the first branch of the encoder-decoder, we incorporate multiple SDTBs to capture intricate features and discern dynamically distributed rain-related factors. 
For the second branch, we leverage multiple CBSMs to supplement the missing information for the first branch of Transformer blocks. 
The output features of two branches are merged at each level by concatenation and point-wise convolution.
Within the same level, encoders and decoders are interconnected through skip-connections, aligning with prior studies~\cite{zamir2022restormer,xiao2022image,wang2022uformer}, to bolster the training process's stability. 
Between each level, we employ pixel-unshuffle and pixel-shuffle operations, serving the purpose of feature down-sampling and up-sampling, respectively. 

\subsection{Spectral-Domain Transformer Block}
\label{sec:SDTB}

At each level of our TransMamba, multiple SDTBs are cascaded and the pipeline of SDTB can be expressed by the following equations:
\begin{align}
    F_{n} &= F_{n-1} + {\rm SBSA} \left[{\rm LayerNorm} \left(F_{n-1}\right)\right],\\
    F_{n} &=  F_{n} + {\rm SEFF} \left[{\rm LayerNorm} \left( F_{n}\right)\right],
\end{align}
where \(LayerNorm\) denotes layer normalization, and \(F_{n}\) represents the feature at the \(n\)-th level. 
It comprises two sub-modules, i.e., SBSA for attaining long-range and local dependency extraction in both the spectral and spatial domains, and SEFF for augmenting frequency-specific information of rain streaks. 
The details of these sub-modules are elaborated in the subsequent sections.

\subsubsection{Spectral Banding Self-Attention Module}
\label{sec:SBSA}
\looseness-1
To better capture dynamically distributed rain-induced degradation, we introduce a spectral banding self-attention (SBSA) module. 
This module encompasses a sequence involving a group of separable convolutions, consisting of a point-wise convolution and a depth-wise convolution, a 2D FFT operation, and a spectral banding attention mechanism.
Separable convolutions are employed to extract rain-related degradation features of Query, Key, and Value. 
This extraction allows the subsequent attention mechanism to weigh the values of feature tokens effectively.
The operation of FFT transforms spatial-domain features into the spectral domain, where each pixel embeds a sinusoidal component of the original spatial features with a specific frequency. 
Due to the decomposition property, modifying a single token in the spectral domain yields a global signal alteration of the original image. 
We thus argue that each spectral-domain token encodes partial dependencies of global knowledge, and the extraction of the tokens enables the capture of long-range information.
By employing the aforementioned process of feature extraction, our SBSA achieves a combination of global and local dynamic feature aggregation. 
This process is articulated as follows:
\begin{equation}
\begin{split}
    &[F_{Q},F_{K},F_{V}] = {\rm Conv_{1\times 1}}(F_n), \\
    &F_{i\in\{Q,K,V\}} = {\rm C2R}(\fft( {\rm Conv^d_{3\times 3}}(F_i) )),
\end{split}
\end{equation}
where \({\rm Conv^d_{3\times 3}}\) is the \(3\times 3\) depth-wise convolution, \({\rm Conv_{1\times 1}}\) represents the point-wise convolution, \(\fft\) denotes the operation of 2D FFT, \({\rm C2R}\) represents the complex-to-real operation achieved by {\ttfamily view\_as\_real} in PyTorch, and \(F_{i\in\{Q,K,V\}}\in \mathbb{R}^{C\times H\times W}\) are the Query, Key, and Value features in the spectral domain. 

\looseness-1
Existing vision Transformers~\cite{xiao2022image,zamir2022restormer,wang2022uformer,chen2023learning,zhao2023comprehensive} typically rely on a fixed range of attention or attention solely along the channel dimension due to computational and memory efficiency compromises. 
However, this fixed setting restricts self-attention from adaptively spanning long ranges to associate desired features.
In contrast, we observe that rain-induced degradation results in repetitive textures encoded within the restricted range of low frequency in the spectral domain, and pixels of different frequencies are better assigned with varying extents of attention. 
Therefore, we propose a spectral banding reorganization (SBR) to categorize spectral-domain elements into bands of different frequencies and allocate varying attentions across them. 
We first gather the tokens according to a mesh index sorting the spectral features from high to low frequency. 
The reordered spectral-domain features are then categorized into band sequences with the same length and concatenated across bands.
For the sake of parallel computing, we restrict that each band contains an identical number of pixels during implementation.
The process of SBR is expressed as
%
\begin{equation}
\begin{split}
    F_{i\in\{Q,K,V\}} = {\rm Reorganize}({\rm gather}(F_i;\mathcal{M})),
\end{split}
\end{equation}
where \({\rm Reorganize}\) represents the operation of reorganizing features from \(\mathbb{R}^{C\times HW}\) to \(\mathbb{R}^{Cb\times HW/b}\), and \({\rm gather}(\cdot; \mathcal{M})\) is the gathering operation following the mesh index of \(\mathcal{M}\). 
The index \(\mathcal{M}\) is pre-computed as a form of mesh grid for a specific shape of \(H\times W\) such that its center items possess small indices indicating higher frequencies while the outer indices indicating lower frequencies are ordered at last.
To extract both long-range and local information, we mix the channel dimension, involving rich information of local representation, and the band dimension, encoding the dynamic-range dependencies. 
The reorganized features are then fed into the self-attention mechanism, and the process can be formulated as below:
\begin{equation}
    \begin{split}
        &F_n = {\rm softmax}\left(\frac{F_QF_K^{\top}} {\sqrt{k}}\right) F_V,
    \end{split}
\end{equation}
where \(k\) is the number of heads. 
After obtaining spectral-domain attentive feature, we output the spatial feature by
\begin{equation*}
     F_n = {\rm Conv_{1\times 1}} \left(\ifft \left({\rm R2C}\left( {\rm Scatter} \left({\rm R} (F_n); \mathcal{M}\right) \right) \right) \right),
\end{equation*}
where \(\ifft\) is the process of 2D inverse FFT, \({\rm R2C}\) represents the real-to-complex operation achieved by {\ttfamily view\_as\_complex} in PyTorch, \({\rm Scatter}\) denotes the token retrieval step based on the index \(\mathcal{M}\), and \({\rm R}\) is the dimension-reshaping operation.

\subsubsection{Spectral Enhanced Feed-Forward Module}
\label{sec:SEFF}

\looseness-1
Previous studies~\cite{zamir2022restormer,xiao2022image,wang2022uformer,chen2023learning} typically rely on single-range or single-scale convolution in the feed-forward network to enhance local context. 
However, these methods often overlook correlations among dynamically distributed rain-induced degradation of different ranges and scales. 
In practice, multi-scale information can be extracted not only by enlarging the kernel size but also by leveraging the dilation mechanism~\cite{yu2016multiscale,yu2017dilated,li2023dilated}. 
Consequently, we integrate two distinct multi-range depth-wise convolution pathways in our feed-forward module. 
We also capitalize on the property that convolution in the spatial domain is equivalent to element-wise product in the spectral domain, leading to the conception of the spectral enhanced feed-forward (SEFF) module. 
In SEFF, a pair of weights and biases is resized and broadcast to act as a biased filter on the spectral-domain features. 
This enables each token, representing a frequency component, to be adaptively enhanced or filtered, such that rain-related and background tokens can be better separated.

Given an input tensor \(F_{n} \in \mathbb{R}^{C\times H\times W}\), we initially employ a point-wise convolution operation to augment the channel dimension by a factor of \(r\). 
Following this augmentation, the expanded tensor is directed into two parallel branches. 
Throughout the feature transformation process, \(3\times 3\) and dilated \(3\times 3\) depth-wise convolutions are employed to enhance the extraction of multi-range information. 
The features on both branches are then transformed into the spectral domain and enhanced by a pair of learnable weight and bias variables, such that the spectral-domain information is enhanced. 
The weight has a predefined size, but can be interpolated to a shape fitting the input feature and acts like an adaptive-frequency-pass filter, remaining background tokens and filtering out rain-related degradation.
Following the gating mechanism~\cite{dauphin2017language}, the activated output of one branch with the receptive field of longer range acts as a gating unit for the other branch. 
Thus, the complete feature fusion process within the SEFF module is formulated as follows:
\begin{equation*}
\begin{split}
    &[F_{1}, F_{2}] = {\rm Conv}_{1\times 1} (F_{n}), \\
    &F_{1} = {\rm Conv^d_{3\times 3}} (F_{1}),\ 
    F_{2} = {\rm Conv^{d,dilated}_{3\times 3}} (F_{2}), \\
    &F_{1} = \mathcal{W}^\uparrow_1\odot \fft(F_{1}) + \mathcal{B}_1,\ F_{2} =\mathcal{W}^\uparrow_2\odot \fft(F_{2}) + \mathcal{B}_2,\\
    &F_{n} = {\rm Conv}_{1\times 1}\left(\left({\rm SiLU}(\ifft(F_{2})) \odot \ifft(F_{1}) \right)\right),
\end{split}
\end{equation*}
where \({\rm Conv^{d,dilated}_{3\times 3}}\) is \(3\times 3\) dilated depth-wise convolution, \(\mathcal{W}^\uparrow_{i\in \{1, 2\}}\in \mathbb{C}^{C\times H\times W}\) represents the interpolated feature map of spectral-domain weight, \(\mathcal{B}_{i\in \{1, 2\}}\in \mathbb{R}^{C\times 1\times 1}\) is the bias feature map, and \({\rm SiLU}\) is the SiLU activation~\cite{elfwing2018sigmoid}.

\subsection{Cascaded Bidirectional SSM Modules}
\label{sec:CBSM}
Inspired by~\cite{zhu2024visionmamba}, we propose leveraging cascaded bi-directional SSM modules to further enhance the sequential coherence. 
The expression of forward direction can be formulated as below,
\begin{equation*}
\begin{split}
    &[F_1, F_2] = {\rm Conv_{1\times 1}}(F_n), \\
    &F_1 = {\rm SSM}(\sigma({\rm Conv1d^d}({\rm SA} ( {\rm Conv^d_{5\times 5}}(F_1) )))), \\
    &F_3 = F_1 \cdot \sigma({\rm CA} ( {\rm Conv^d_{5\times 5}}(F_2) )),\\
\end{split}
\end{equation*}
where \(\rm SA\) and \(\rm CA\) denote the spatial attention and channel attention~\cite{woo2018cbam} respectively.
The expression of backward direction can be formulated as below,
\begin{equation*}
\begin{split}
&[F_4, F_5] = {\rm Conv_{1\times 1}}({\rm Flip}(F_3)), \\
    &F_4 = {\rm SSM}(\sigma({\rm Conv1d^d}({\rm SA} ( {\rm Conv^d_{5\times 5}}(F_4) )))), \\
    &F_o = F_4 \cdot \sigma({\rm CA} ( {\rm Conv^d_{5\times 5}}(F_5) )),
\end{split}
\end{equation*}
where \(\rm Flip\) is the flip operation along channel dimension. 
With the channel-flipping operation, the convolution and SSM processes are reversed from the forward direction to the backward direction. 
Cascading the forward- and backward-directional CBSMs forms the bi-directional SSM module.
The bi-directional strategy can alleviate the forgetting of long range feature of SSM modules, better enhancing the long range information extraction.

\subsection{Reconstruction Loss and Coherence Loss}\label{sec:corloss}

To supervise the training of our TransMamba, we employ \(L_1\) norm between the derained image \(\tilde B\) and clean background \(B\) as the reconstruction loss, which is expressed as follows:
\begin{equation}
    \mathcal{L}_{rec} = \left\|\tilde B - B\right\|_1.
\end{equation}
%


\begin{table*}[tb]
\footnotesize
\renewcommand\arraystretch{1.3}
  \centering
  \caption{Quantitative Comparison among our SDBFormer and the existing methods on both synthetic and real rain datasets. The bottom half of the table contains the Transformer-based methods, while the upper half contains the others. The values in \textbf{bold} and \underline{underlined} are the best and the second best results.}
    \begin{tabular}{llccccccccccc}
    \toprule
    \multirow{2.5}{*}{Type}  &  \multirow{2.5}{*}{Methods} & Datasets & \multicolumn{2}{c}{Rain200H} & \multicolumn{2}{c}{Rain200L} & \multicolumn{2}{c}{DID-Data} & \multicolumn{2}{c}{DDN-Data} & \multicolumn{2}{c}{SPA-Data} \\
    \cmidrule{3-13} &  & Venue & PSNR  & SSIM  & PSNR  & SSIM  & PSNR  & SSIM  & PSNR  & SSIM  & PSNR  & SSIM \\
    \midrule
    \multirow{2}{*}{Prior} 
    & DSC~\cite{luo2015removing} & \textit{ICCV'15}  & 14.73 & 0.3815 & 27.16 & 0.8663 & 24.24 & 0.8279 & 27.31 & 0.8373 & 34.95 & 0.9416 \\
    & GMM~\cite{li2016rain}  & \textit{CVPR'16} & 14.50 & 0.4164 & 28.66 & 0.8652 & 25.81 & 0.8344 & 27.55 & 0.8479 & 34.30 & 0.9428 \\
    \midrule
    \multirow{8}{*}{CNN}
    & DDN~\cite{fu2017removing}  & \textit{CVPR'17}  & 26.05 & 0.8056 & 34.68 & 0.9671 & 30.97 & 0.9116 & 30.00 & 0.9041 & 36.16 & 0.9457 \\
    & RESCAN~\cite{li2018recurrent} & \textit{ECCV'18} & 26.75 & 0.8353 & 36.09 & 0.9697 & 33.38 & 0.9417 & 31.94 & 0.9345 & 38.11 & 0.9707 \\
    & PReNet~\cite{ren2019progressive} & \textit{CVPR'19} & 29.04 & 0.8991 & 37.80 & 0.9814 & 33.17 & 0.9481 & 32.60 & 0.9459 & 40.16 & 0.9816 \\
    & MSPFN~\cite{jiang2020multi} & \textit{CVPR'20} & 29.36 & 0.9034 & 38.58 & 0.9827 & 33.72 & 0.9550 & 32.99 & 0.9333 & 43.43 & 0.9843 \\
    & RCDNet~\cite{wang2020model} & \textit{CVPR'20} & 30.24 & 0.9048 & 39.17 & 0.9885 & 34.08 & 0.9532 & 33.04 & 0.9472 & 43.36 & 0.9831 \\
    & MPRNet~\cite{zamir2021multi} & \textit{CVPR'21} & 30.67 & 0.9110 & 39.47 & 0.9825 & 33.99 & 0.9590 & 33.10 & 0.9347 & 43.64 & 0.9844 \\
    & DualGCN~\cite{Fu2021RainSR} & \textit{AAAI'21} & 31.15 & 0.9125 & 40.73 & 0.9886 & 34.37 & 0.9620 & 33.01 & 0.9489 & 44.18 & 0.9902 \\
    & SPDNet~\cite{yi2021Structure} & \textit{ICCV'21} & 31.28 & 0.9207 & 40.50 & 0.9875 & 34.57 & 0.9560 & 33.15 & 0.9457 & 43.20 & 0.9871 \\
    \midrule
    \multirow{6}{*}{Transformer} 
    & Uformer~\cite{wang2022uformer} & \textit{CVPR'22} & 30.80 & 0.9105 & 40.20 & 0.9860 & 35.02 & 0.9621 & 33.95 & 0.9545 & 46.13 & 0.9913 \\
    & Restormer~\cite{zamir2022restormer} & \textit{CVPR'22} & 32.00 & 0.9329 & 40.99 & 0.9890 & 35.29 & 0.9641 & 34.20 & 0.9571 & 47.98 & 0.9921 \\
    & IDT~\cite{xiao2022image} & {\textit{PAMI'22}}  & 32.10 & 0.9344 & 40.74 & 0.9884 & 34.89 & 0.9623 & 33.84 & 0.9549 & 47.35 & 0.9930 \\
    & DRSformer~\cite{chen2023learning} & \textit{CVPR'23} & 32.17 & 0.9326 & 41.23 & 0.9894 & 35.35 & 0.9646 & 34.35 & 0.9588 & 48.54 & 0.9924 \\
    & UDR-S$^2$Former~\cite{chen2023sparse} & \textit{ICCV'23} & 32.59 & 0.9374 & 40.96 & 0.9892 & 35.29 & 0.9628 & 34.41 & 0.9573 & 48.57 & 0.9917 \\
    & NeRD-Rain~\cite{NeRD-Rain} & \textit{CVPR'24}& 32.40 & 0.9373 & 41.71 & \underline{0.9903} & \underline{35.53} & \textbf{0.9659} & \underline{34.45} & \underline{0.9596} & \underline{49.58} & 0.9940 \\
    \midrule
    \multirow{1}{*}{Mamba} 
    & DFSSM~\cite{yamashita2024image} & ArXiv'24 & \underline{32.90} & \underline{0.9394}  & \underline{41.73} & 0.9900 & - & - & - & - & 48.83 & \underline{0.9944} \\
    \midrule
    Hybrid
    & Ours & - & \textbf{32.96} & \textbf{0.9409} & \textbf{41.92} & \textbf{0.9938} & \textbf{35.63} & \underline{0.9657} & \textbf{34.72} & \textbf{0.9603} & \textbf{49.72} & \textbf{0.9968} \\
    \bottomrule
    \end{tabular}%
  \label{tab:result}%
\end{table*}%

\begin{figure*}[tb]
  \centering
     \subfloat[Input]{
  \begin{minipage}{0.116\linewidth}
    \centering
    \includegraphics[trim=0mm 60mm 0mm 0mm,clip,width=1\linewidth]{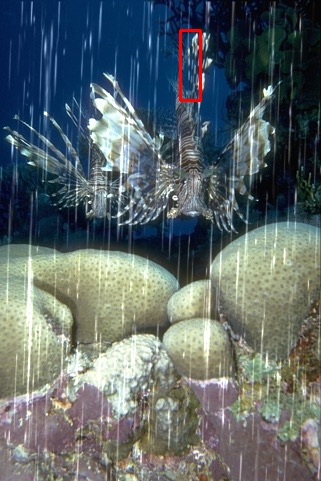}
    \hfill
    \includegraphics[width=1\linewidth]{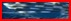}
    \hfill
    \includegraphics[width=1\linewidth]{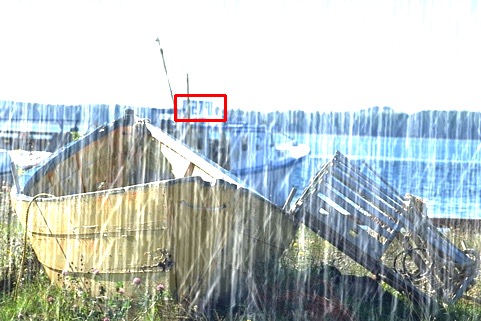}
    \hfill
    \includegraphics[width=1\linewidth]{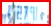}
    \end{minipage}
    }
  \hspace{-2.95mm}
     \subfloat[\cite{zamir2022restormer}]{
  \begin{minipage}{0.116\linewidth}
    \centering
    \includegraphics[trim=0mm 60mm 0mm 0mm,clip,width=1\linewidth]{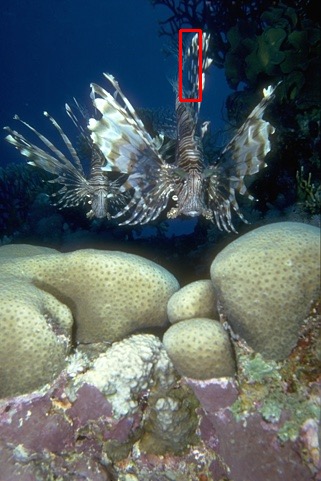}
    \hfill
    \includegraphics[width=1\linewidth]{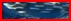}
    \hfill
    \includegraphics[width=1\linewidth]{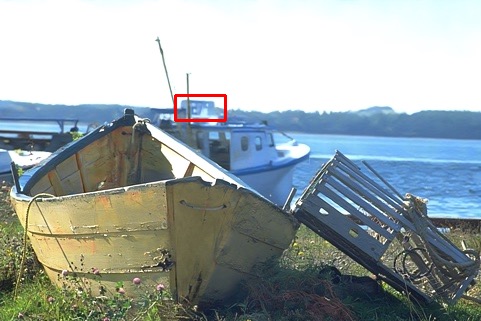}
    \hfill
    \includegraphics[width=1\linewidth]{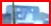}
    \end{minipage}
    }
  \hspace{-2.95mm}
     \subfloat[\cite{xiao2022image}]{
  \begin{minipage}{0.116\linewidth}
    \centering
    \includegraphics[trim=0mm 60mm 0mm 0mm,clip,width=1\linewidth]{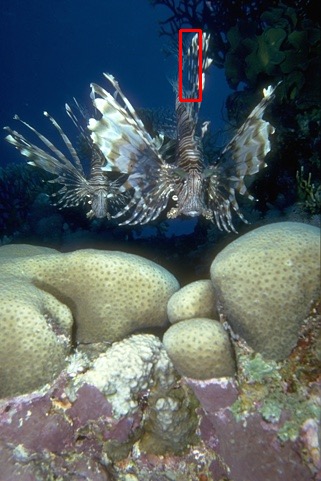}
    \hfill
    \includegraphics[width=1\linewidth]{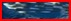}
    \hfill
    \includegraphics[width=1\linewidth]{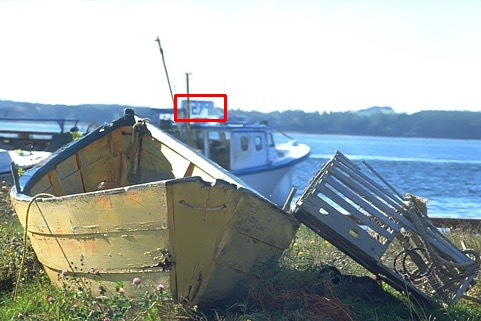}
    \hfill
    \includegraphics[width=1\linewidth]{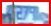}
    \end{minipage}
    }
  \hspace{-2.95mm}
     \subfloat[\cite{chen2023learning}]{
  \begin{minipage}{0.116\linewidth}
    \centering
    \includegraphics[trim=0mm 60mm 0mm 0mm,clip,width=1\linewidth]{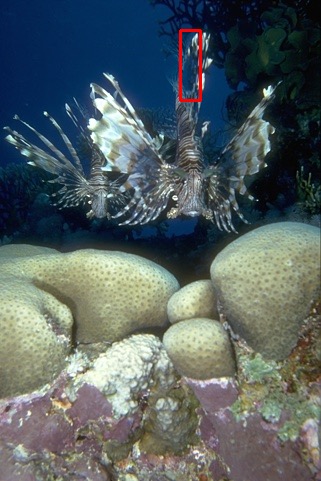}
    \hfill
    \includegraphics[width=1\linewidth]{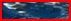}
    \hfill
    \includegraphics[width=1\linewidth]{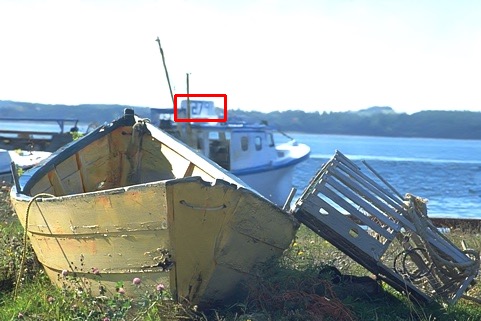}
    \hfill
    \includegraphics[width=1\linewidth]{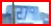}
    \end{minipage}
    }
  \hspace{-2.95mm}
     \subfloat[\cite{chen2023sparse}]{
  \begin{minipage}{0.116\linewidth}
    \centering
    \includegraphics[trim=0mm 60mm 0mm 0mm,clip,width=1\linewidth]{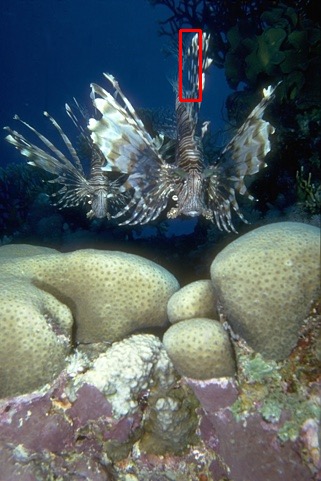}
    \hfill
    \includegraphics[width=1\linewidth]{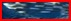}
    \hfill
    \includegraphics[width=1\linewidth]{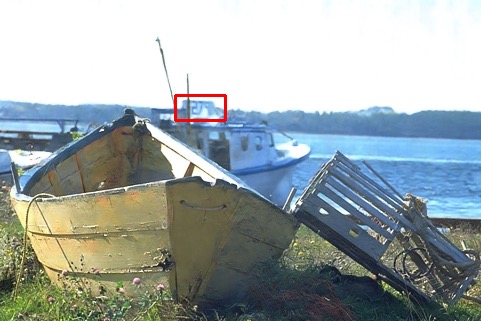}
    \hfill
    \includegraphics[width=1\linewidth]{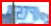}
    \end{minipage}
    }
  \hspace{-2.95mm}
     \subfloat[\cite{NeRD-Rain}]{
  \begin{minipage}{0.116\linewidth}
    \centering
    \includegraphics[trim=0mm 60mm 0mm 0mm,clip,width=1\linewidth]{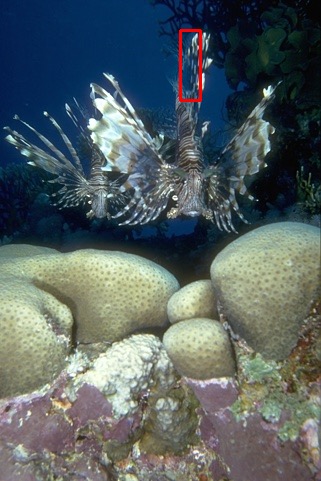}
    \hfill
    \includegraphics[width=1\linewidth]{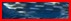}
    \hfill
    \includegraphics[width=1\linewidth]{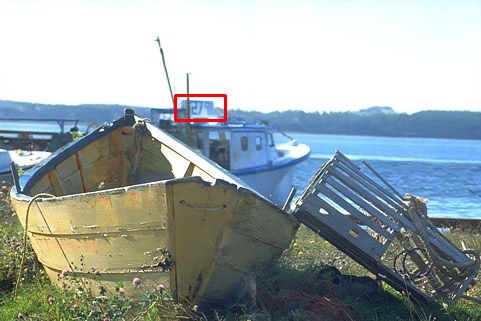}
    \hfill
    \includegraphics[width=1\linewidth]{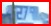}
    \end{minipage}
    }
  \hspace{-2.95mm}
     \subfloat[Ours]{
  \begin{minipage}{0.116\linewidth}
    \centering
    \includegraphics[trim=0mm 60mm 0mm 0mm,clip,width=1\linewidth]{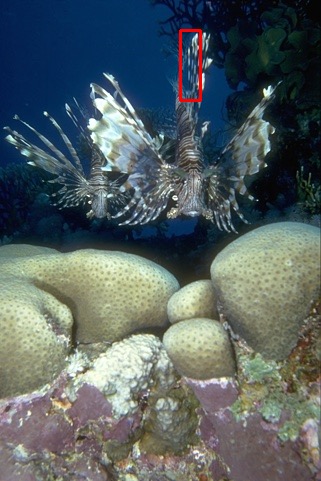}
    \hfill
    \includegraphics[width=1\linewidth]{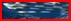}
    \hfill
    \includegraphics[width=1\linewidth]{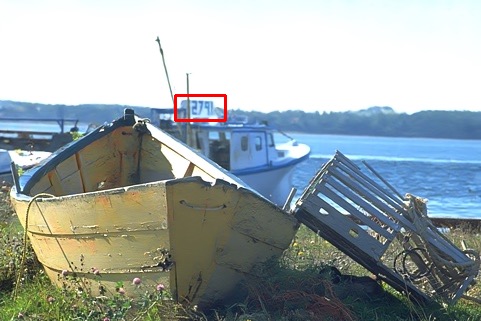}
    \hfill
    \includegraphics[width=1\linewidth]{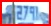}
    \end{minipage}
    }
  \hspace{-2.95mm}
     \subfloat[Clean]{
  \begin{minipage}{0.116\linewidth}
    \centering
    \includegraphics[trim=0mm 60mm 0mm 0mm,clip,width=1\linewidth]{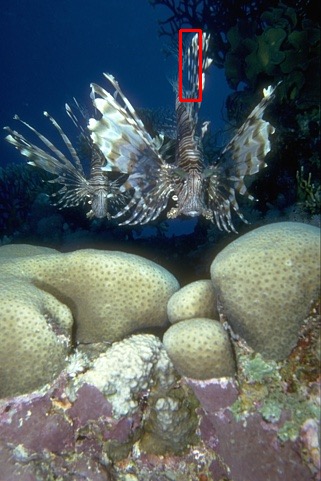}
    \hfill
    \includegraphics[width=1\linewidth]{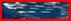}
    \hfill
    \includegraphics[width=1\linewidth]{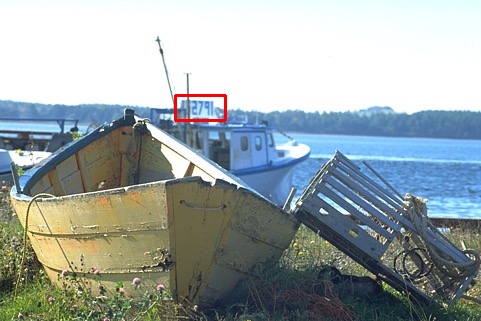}
    \hfill
    \includegraphics[width=1\linewidth]{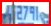}
    \end{minipage}
    }
  \caption{Visual comparisons of deraining on Rain200L (1st row) and Rain200H (2nd row)~\cite{yang2017deep}. The sample from (b) to (f) are Restormer~\cite{zamir2022restormer}, IDT~\cite{xiao2022image}, DRSformer~\cite{chen2023learning}, UDR-S$^2$Former~\cite{chen2023sparse}, and NeRD-Rain~\cite{NeRD-Rain}, respectively. Please zoom in for a better view.}
  \label{fig:rain200lh}
\end{figure*}

Furthermore, we notice that the \(\mathcal{L}_{rec}\) only regulates the pixel-level similarity between the derained image and the ground-truth, while disregarding the global coherence between them from the perspective of signal. 
The innate relationships of signal components in the spectral domain within the image are disrupted by the patterns of rain degradation. 
By emulating the original signal relationships within the ground-truth, we compel the signal representation of degraded images to recover to their original sinusoidal signals of the clean background.
Consequently, we introduce the coherence~\cite{bendat2011random} between image spectra as a means to regulate the comprehensive linear relationship, expressed as follows:
\begin{equation}\label{eq::cohloss}
\small
    \mathcal{G}\left( \tilde B, B\right) = \frac{ \left\| \fft(\tilde B)\overline{\fft( B)}\right\|^2_1}{ \left\| \fft(\tilde B)\overline{\fft( \tilde B)}\right\|_1 \left\| \fft(B)\overline{\fft( B)}\right\|_1 },
\end{equation}
where \(\overline F\) denotes the conjugate component of a signal \(F\), \(\left\| \fft(F)\overline{\fft(F)}\right\|_1\) is the spectral density of a signal \(F\), and the numerator \(\left\| \fft(\tilde B)\overline{\fft( B)}\right\|^2_1\) represents the squared cross-spectral density between the derained and clean image signals. 
The value of Eq.~\ref{eq::cohloss} falls within the range of \([0,1]\). 
When two image signals exhibit irrelevant coherence, Eq.~\ref{eq::cohloss} reaches a value of \(0\), while in the case of perfectly linear coherence, it obtains a value of \(1\). Hence, we formulate the spectral coherence loss as follows:
\begin{equation}
    \mathcal{L}_{coh} = 1-\sqrt{\mathcal{G}\left( \tilde B, B\right)}.
\end{equation}
The final loss function is thus expressed as:
\begin{equation}
    \mathcal{L} = \mathcal{L}_{rec} +\alpha \mathcal{L}_{coh},
\end{equation}
where \(\alpha\) is the weight of spectral coherence loss.

\begin{figure*}[tb]
  \centering
    \subfloat[Input]{
  \begin{minipage}{0.106\linewidth}
    \centering
    \includegraphics[width=1\linewidth]{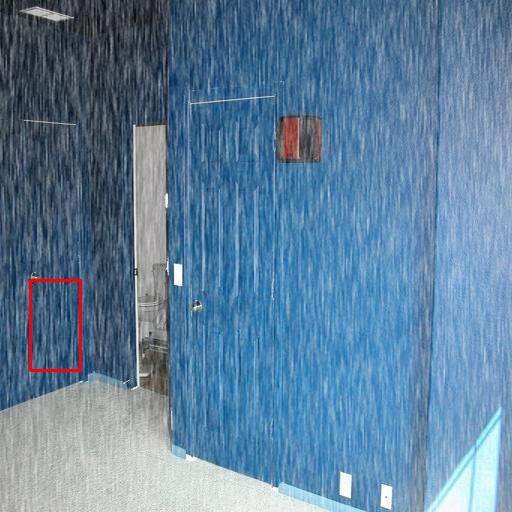}
    \hfill
    \includegraphics[width=1\linewidth]{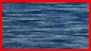}
    \hfill
    \includegraphics[width=1\linewidth]{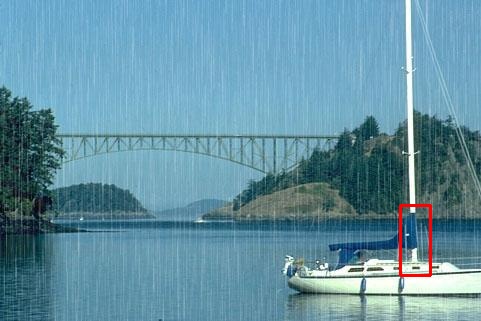}
    \hfill
    \includegraphics[width=1\linewidth]{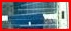}
    \end{minipage}
    }
  \hspace{-1.5mm}
    \subfloat[\cite{zamir2022restormer}]{
  \begin{minipage}{0.106\linewidth}
    \centering
    \includegraphics[width=1\linewidth]{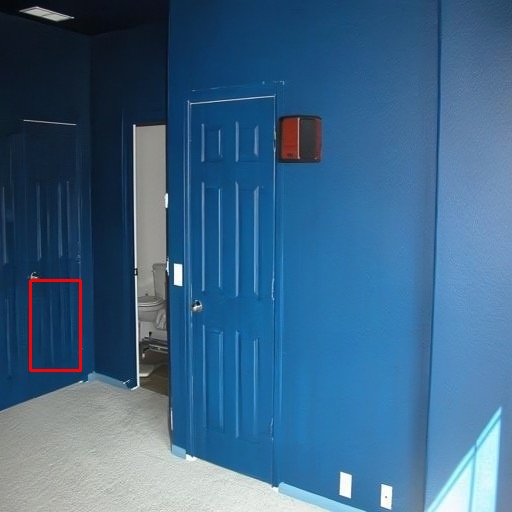}
    \hfill
    \includegraphics[width=1\linewidth]{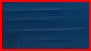}
    \hfill
    \includegraphics[width=1\linewidth]{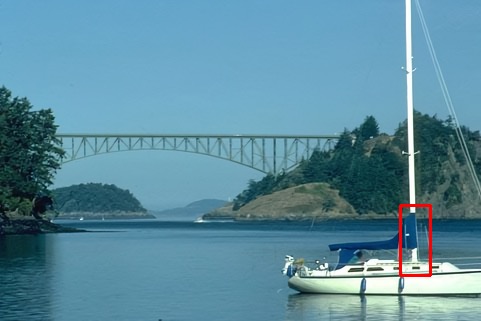}
    \hfill
    \includegraphics[width=1\linewidth]{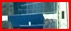}
    \end{minipage}
    }
  \hspace{-1.5mm}
    \subfloat[\cite{xiao2022image}]{
  \begin{minipage}{0.106\linewidth}
    \centering
    \includegraphics[width=1\linewidth]{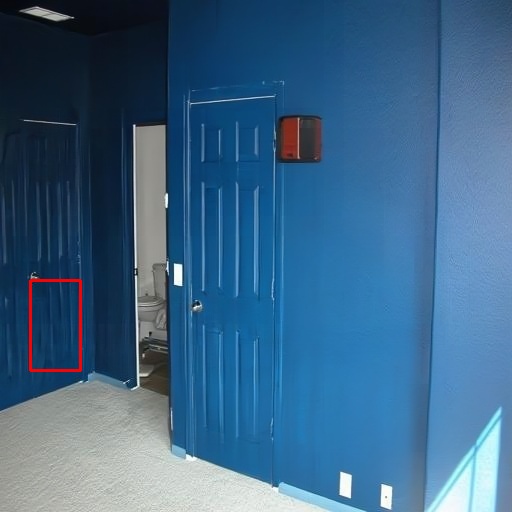}
    \hfill
    \includegraphics[width=1\linewidth]{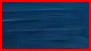}
    \hfill
    \includegraphics[width=1\linewidth]{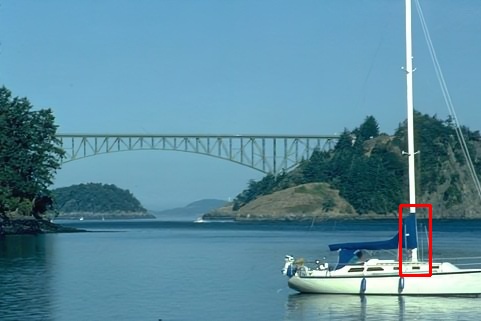}
    \hfill
    \includegraphics[width=1\linewidth]{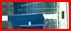}
    \end{minipage}
    }
  \hspace{-1.5mm}
    \subfloat[\cite{chen2023learning}]{
  \begin{minipage}{0.106\linewidth}
    \centering
    \includegraphics[width=1\linewidth]{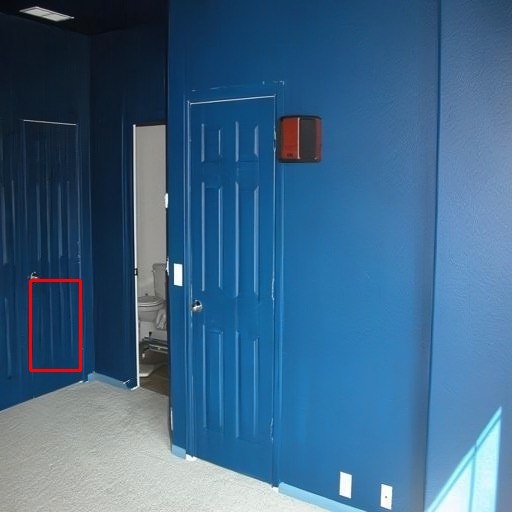}
    \hfill
    \includegraphics[width=1\linewidth]{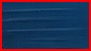}
    \hfill
    \includegraphics[width=1\linewidth]{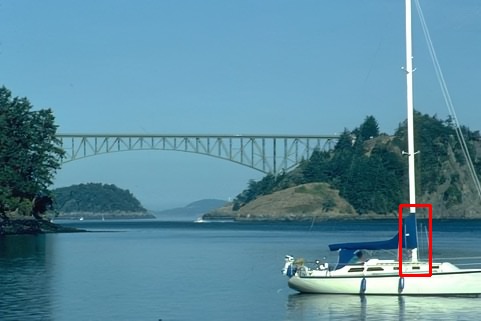}
    \hfill
    \includegraphics[width=1\linewidth]{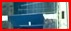}
    \end{minipage}
    }
  \hspace{-1.5mm}
    \subfloat[\cite{chen2023sparse}]{
  \begin{minipage}{0.106\linewidth}
    \centering
    \includegraphics[width=1\linewidth]{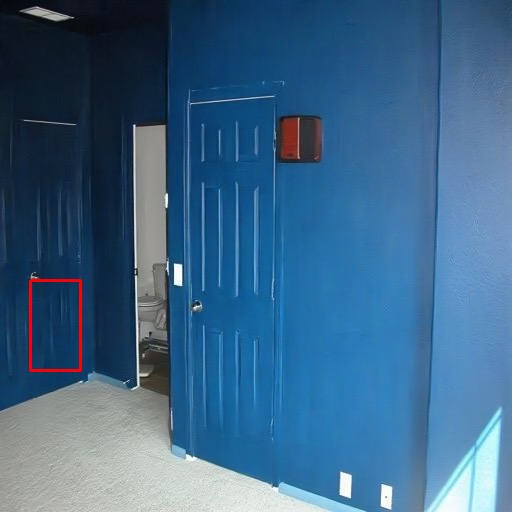}
    \hfill
    \includegraphics[width=1\linewidth]{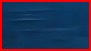}
    \hfill
    \includegraphics[width=1\linewidth]{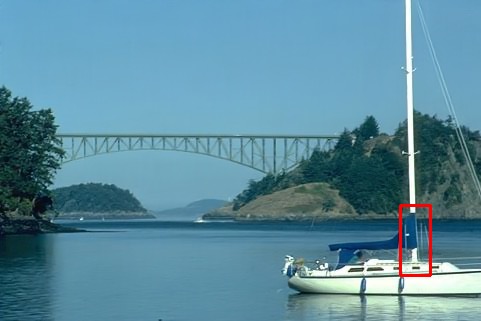}
    \hfill
    \includegraphics[width=1\linewidth]{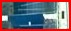}
    \end{minipage}
    }
  \hspace{-1.5mm}
    \subfloat[\cite{NeRD-Rain}]{
  \begin{minipage}{0.106\linewidth}
    \centering
    \includegraphics[width=1\linewidth]{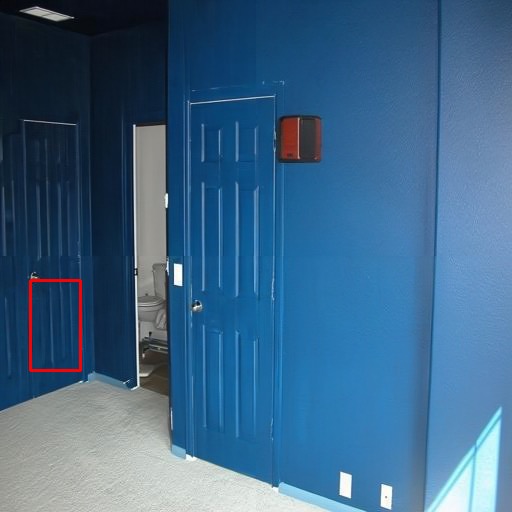}
    \hfill
    \includegraphics[width=1\linewidth]{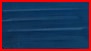}
    \hfill
    \includegraphics[width=1\linewidth]{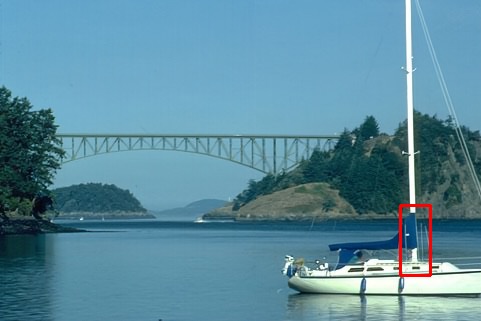}
    \hfill
    \includegraphics[width=1\linewidth]{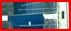}
    \end{minipage}
    }
  \hspace{-1.5mm}
    \subfloat[Ours]{
  \begin{minipage}{0.106\linewidth}
    \centering
    \includegraphics[width=1\linewidth]{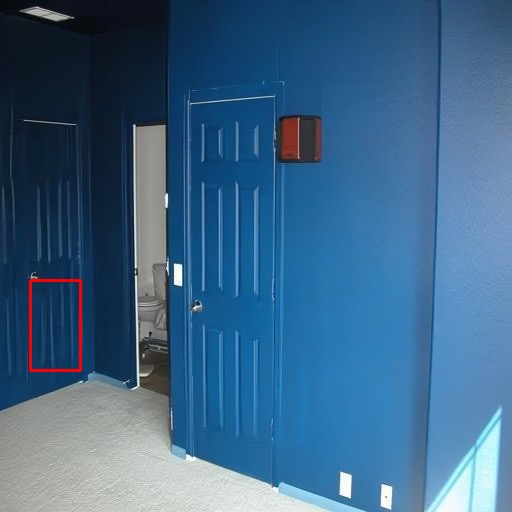}
    \hfill
    \includegraphics[width=1\linewidth]{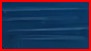}
    \hfill
    \includegraphics[width=1\linewidth]{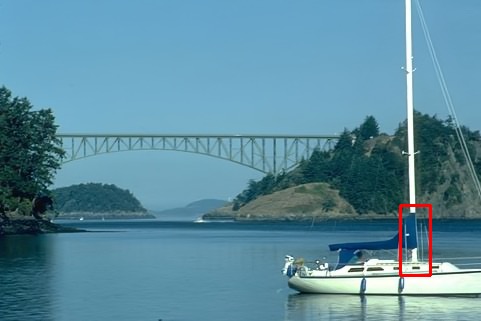}
    \hfill
    \includegraphics[width=1\linewidth]{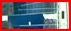}
    \end{minipage}
    }
  \hspace{-1.5mm}
    \subfloat[Clean]{
  \begin{minipage}{0.106\linewidth}
    \centering
    \includegraphics[width=1\linewidth]{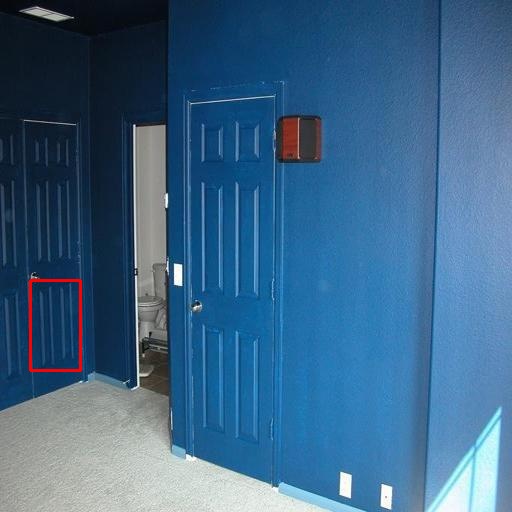}
    \hfill
    \includegraphics[width=1\linewidth]{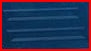}
    \hfill
    \includegraphics[width=1\linewidth]{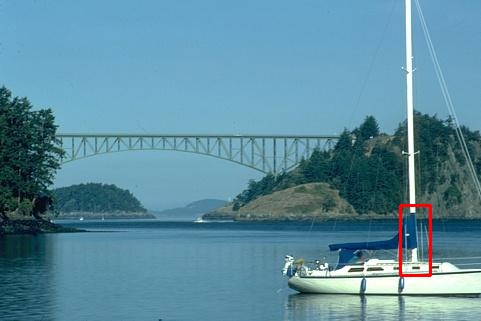}
    \hfill
    \includegraphics[width=1\linewidth]{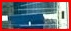}
    \end{minipage}
    }
  \caption{Visual comparisons of deraining on DID-Data~\cite{zhang2018density} (1st row) and DDN-Data~\cite{fu2017removing} (2nd row). The sample from (b) to (f) are Restormer~\cite{zamir2022restormer}, IDT~\cite{xiao2022image}, DRSformer~\cite{chen2023learning}, UDR-S$^2$Former~\cite{chen2023sparse}, and NeRD-Rain~\cite{NeRD-Rain}, respectively. Please zoom in for a better view.}
  \label{fig:did&ddn}
\end{figure*}
\begin{figure*}[tb]
  \centering
    \subfloat[Input]{
  \begin{minipage}{0.120\linewidth}
    \centering
    \includegraphics[width=1\linewidth]{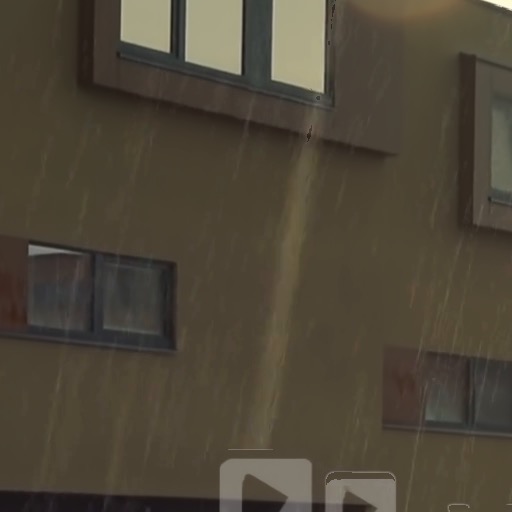}
    \hfill
    \includegraphics[width=1\linewidth]{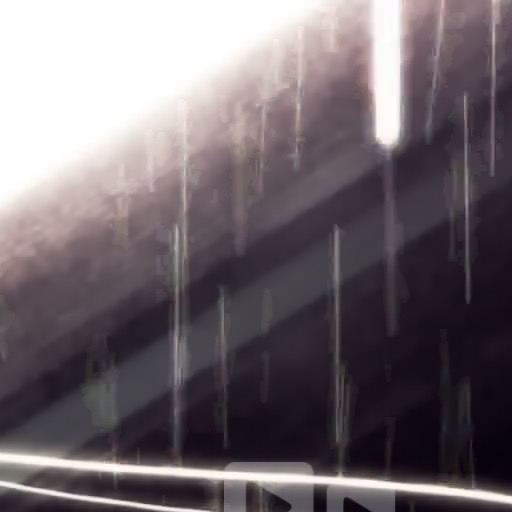}
    \end{minipage}
    }
  \hspace{-3.5mm}
    \subfloat[\cite{yi2021Structure}]{
  \begin{minipage}{0.120\linewidth}
    \centering
    \includegraphics[width=1\linewidth]{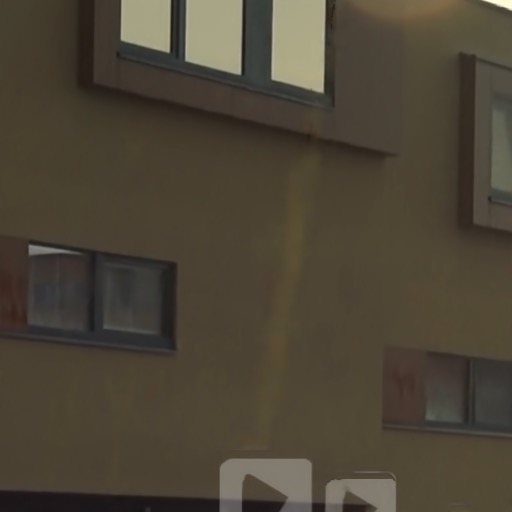}
    \hfill
    \includegraphics[width=1\linewidth]{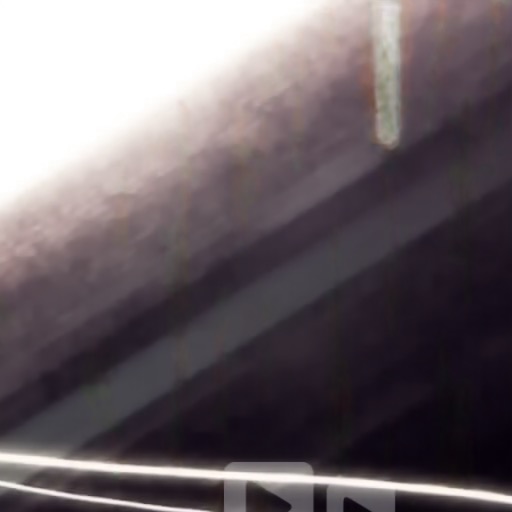}
    \end{minipage}
    }
  \hspace{-3.5mm}
    \subfloat[\cite{zamir2022restormer}]{
  \begin{minipage}{0.120\linewidth}
    \centering
    \includegraphics[width=1\linewidth]{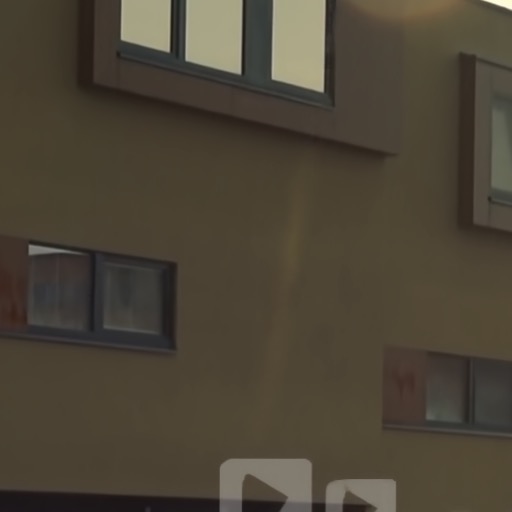}
    \hfill
    \includegraphics[width=1\linewidth]{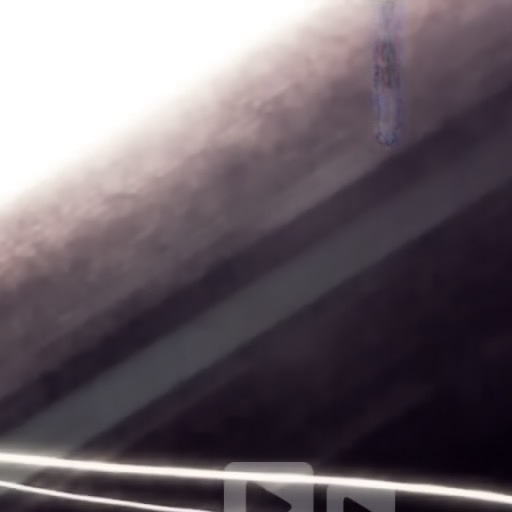}
    \end{minipage}
    }
  \hspace{-3.5mm}
    \subfloat[\cite{xiao2022image}]{
  \begin{minipage}{0.120\linewidth}
    \centering
    \includegraphics[width=1\linewidth]{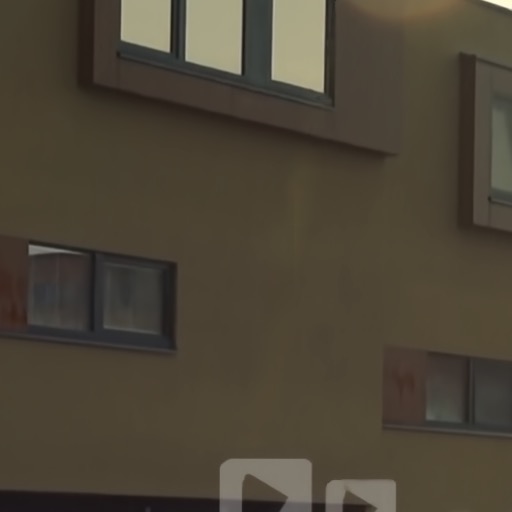}
    \hfill
    \includegraphics[width=1\linewidth]{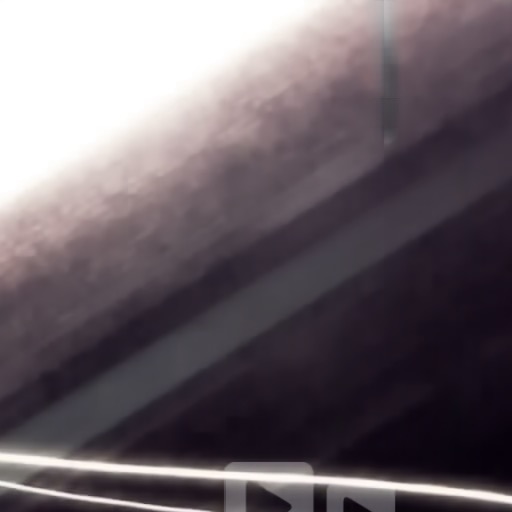}
    \end{minipage}
    }
  \hspace{-3.5mm}
    \subfloat[\cite{chen2023learning}]{
  \begin{minipage}{0.120\linewidth}
    \centering
    \includegraphics[width=1\linewidth]{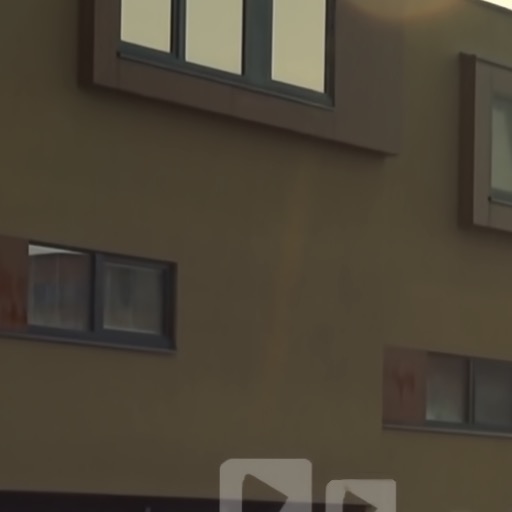}
    \hfill
    \includegraphics[width=1\linewidth]{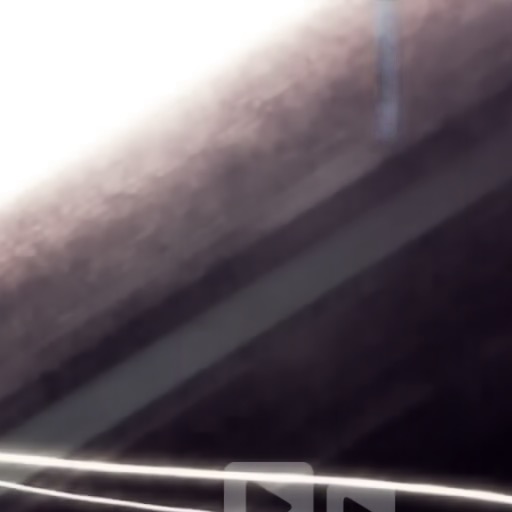}
    \end{minipage}
    }
  \hspace{-3.5mm}
    \subfloat[\cite{chen2023sparse}]{
  \begin{minipage}{0.120\linewidth}
    \centering
    \includegraphics[width=1\linewidth]{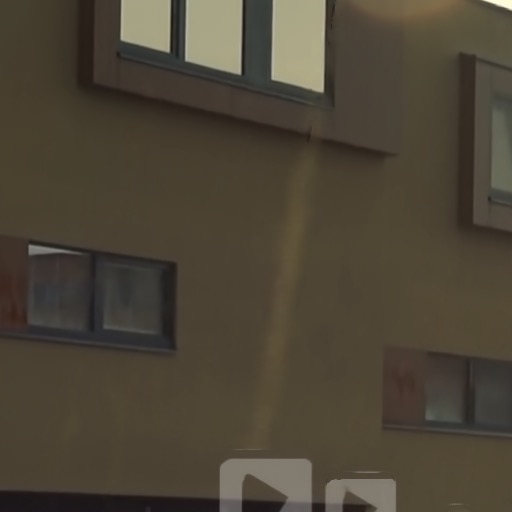}
    \hfill
    \includegraphics[width=1\linewidth]{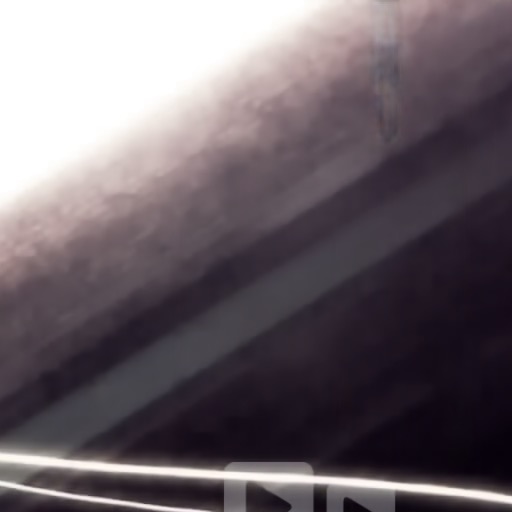}
    \end{minipage}
    }
  \hspace{-3.5mm}
    \subfloat[Ours]{
  \begin{minipage}{0.120\linewidth}
    \centering
    \includegraphics[width=1\linewidth]{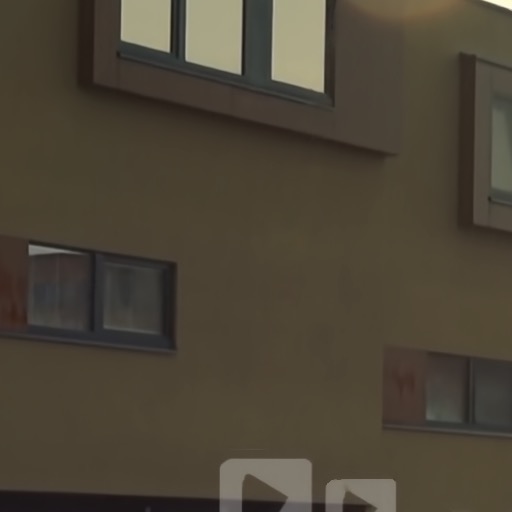}
    \hfill
    \includegraphics[width=1\linewidth]{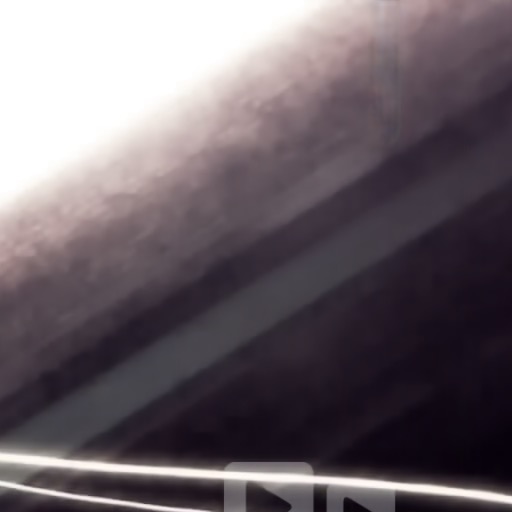}
    \end{minipage}
    }
  \hspace{-3.5mm}
    \subfloat[Clean]{
  \begin{minipage}{0.120\linewidth}
    \centering
    \includegraphics[width=1\linewidth]{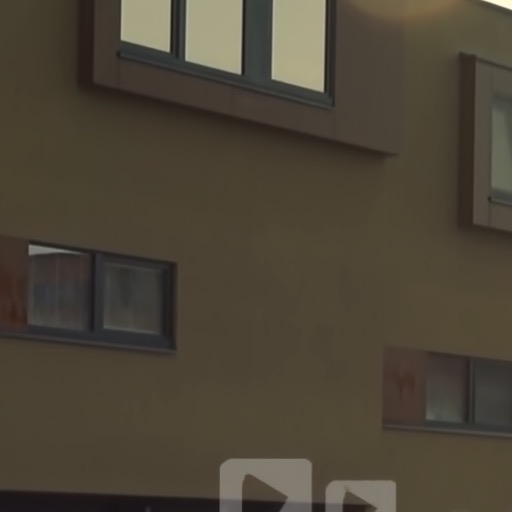}
    \hfill
    \includegraphics[width=1\linewidth]{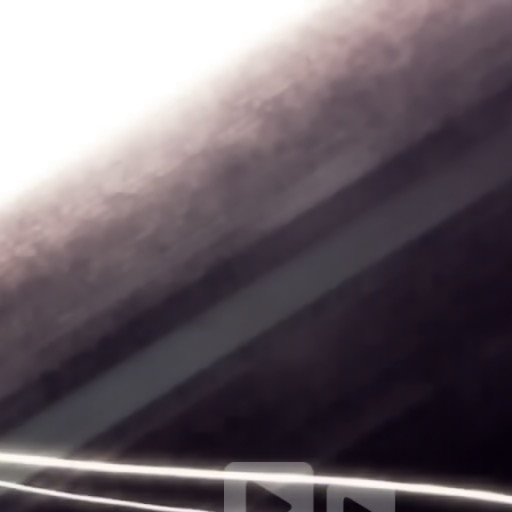}
    \end{minipage}
    }
  \caption{Visual comparisons of deraining on two samples from SPA-Data~\cite{wang2019spatial}. The sample from (b) to (f) are SPDNet~\cite{yi2021Structure}, Restormer~\cite{zamir2022restormer}, IDT~\cite{xiao2022image}, DRSformer~\cite{chen2023learning}, and UDR-S$^2$Former~\cite{chen2023sparse}, respectively. Please zoom in for a better view.}
  \label{fig:spa}
\end{figure*}

\section{Experiment}

\subsection{Experimental Setting}

\paragraph{Implementation Details.} 
In our model, the numbers of SDTBs \(\rm N_{i\in\{1,2,3,4\}}\) and CMSMs \(\rm L_{i\in\{1,2,3,4\}}\) in Figure~\ref{fig:structure} are set to $(1, 3, 4, 4)$ respectively.
The numbers of attention heads for SDTBs at four levels are set to $(1, 2, 4, 8)$. 
The initial channel $C_f$ is set to 36. 
Regarding SBSA, we set \(b=2\) for the number of bands based on the ablation studies in Section~\ref{sec:ablation}. 
In terms of SEFF, the channel expansion factor \(r\) is set to $2.667$ and the predefined weight has the size of \(48\times 48\). 
During training, we use the AdamW optimizer~\cite{loshchilov2019decoupled} with a progressive training~\cite{zamir2022restormer} of an initial batch size $8$ and initial patch size $128$. 
The total number of iterations is $300,000$. 
The initial learning rate is set to \(3 \times 10^{-4}\) for the first $92,000$ iterations, and then reduced to \(1 \times 10^{-6}\) for the remaining $208,000$ iterations using the cosine annealing scheme~\cite{loshchilov2017sgdr}. 
For data augmentation, we use random vertical and horizontal flips. 
The experiments are all implemented on NVIDIA Tesla V100 GPU using PyTorch.

\paragraph{Benchmarks.}
We conduct experiments on five commonly used deraining datasets, namely 
Rain200H~\cite{yang2017deep}, 
Rain200L~\cite{yang2017deep}, 
DID-Data~\cite{zhang2018density}, 
DDN-Data~\cite{fu2017removing} and 
SPA-Data~\cite{wang2019spatial}.
Both Rain200H and Rain200L are introduced by Yang \textit{et al}.~\cite{yang2017deep} and represent two benchmarks with images featuring heavy and light rain streaks, respectively. 
Each of them comprises 1800 synthetic training pairs and 200 test images.
DID-Data~\cite{fu2017removing} consists of 12000 synthetic training pairs characterized by three distinct rain-density levels, alongside 1200 pairs of test images. 
DDN-Data~\cite{fu2017removing} is created by synthesizing 14000 pairs of rainy data based on 1000 images drawn from UCID~\cite{schaefer2003ucid}, BSD~\cite{arbelaez2010contour}, and Google-searched images.
SPA-Data~\cite{wang2019spatial}, is a large-scale real-world rainy dataset, encompassing a vast array of 638,464 rainy/clean image patches for training and 1000 image pairs for testing.
We also utilize Internet-Data~\cite{wang2019spatial}, including 147 real-world rainy images, for qualitative comparison.

\paragraph{Baselines.}
We compare our TransMamba with 
two optimization-based models, i.e., 
DSC~\cite{luo2015removing} and GMM~\cite{li2016rain}, 
eight CNN methods including DDN~\cite{fu2017removing}, 
RESCAN~\cite{li2018recurrent}, 
PReNet~\cite{ren2019progressive}, 
MSPFN~\cite{jiang2020multi}, 
RCDNet~\cite{wang2020model}, 
MPRNet~\cite{zamir2021multi}, 
DualGCN~\cite{Fu2021RainSR} and SPDNet~\cite{yi2021Structure}, 
and five Transformer-based approaches including Uformer~\cite{wang2022uformer}, 
Restormer~\cite{zamir2022restormer}, 
IDT~\cite{xiao2022image}, 
DRSFormer~\cite{chen2023learning}, UDR-S$^2$Former~\cite{chen2023sparse}, and NeRD-Rain~\cite{NeRD-Rain}.

\begin{figure*}[tb]
  \centering
    \subfloat[Input]{
  \begin{minipage}{0.140\linewidth}
    \centering
    \includegraphics[width=1\linewidth]{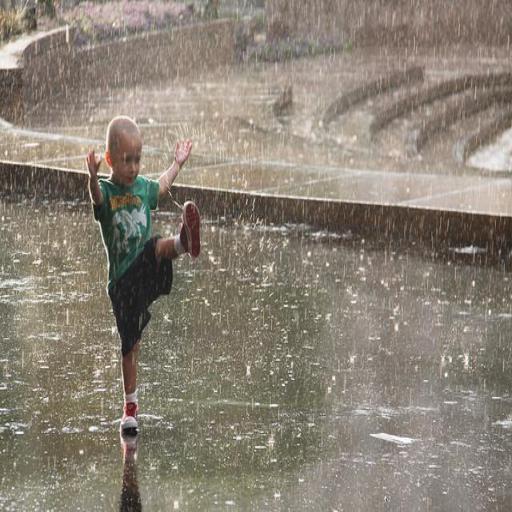}
    \hfill
    \includegraphics[width=1\linewidth]{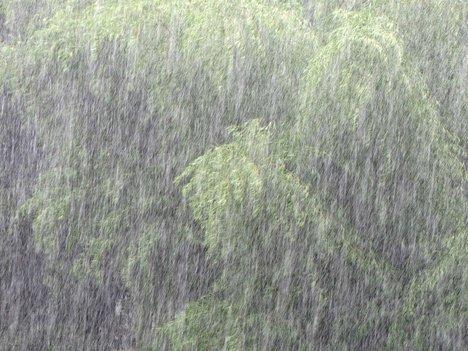}
    \end{minipage}
    }
  \hspace{-3.5mm}
    \subfloat[\cite{zamir2022restormer}]{
  \begin{minipage}{0.140\linewidth}
    \centering
    \includegraphics[width=1\linewidth]{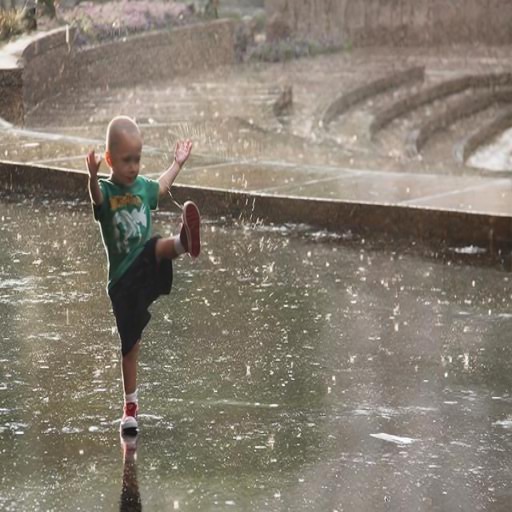}
    \hfill
    \includegraphics[width=1\linewidth]{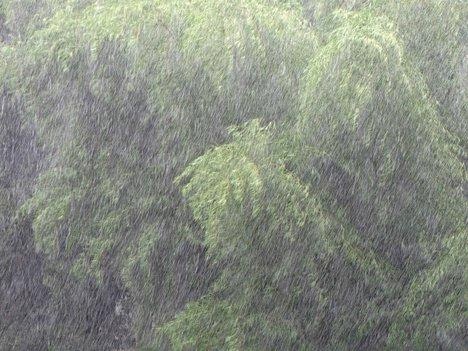}
    \end{minipage}
    }
  \hspace{-3.5mm}
    \subfloat[\cite{xiao2022image}]{
  \begin{minipage}{0.140\linewidth}
    \centering
    \includegraphics[width=1\linewidth]{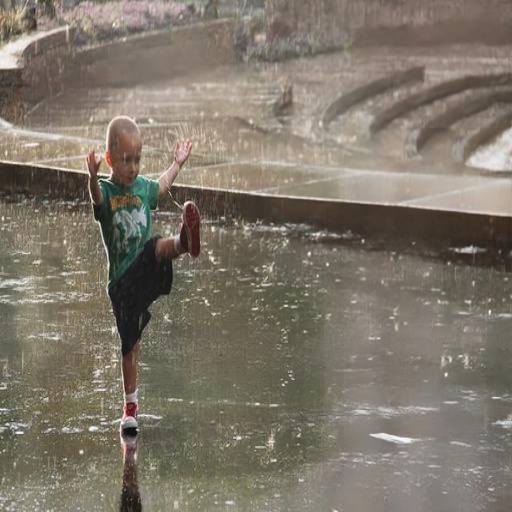}
    \hfill
    \includegraphics[width=1\linewidth]{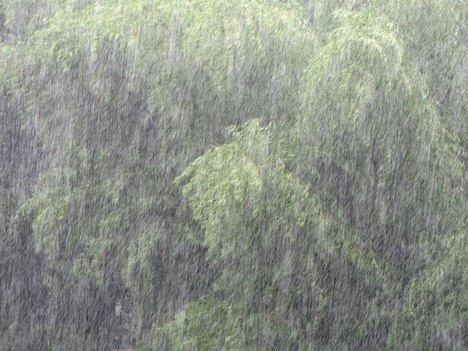}
    \end{minipage}
    }
  \hspace{-3.5mm}
    \subfloat[\cite{chen2023learning}]{
  \begin{minipage}{0.140\linewidth}
    \centering
    \includegraphics[width=1\linewidth]{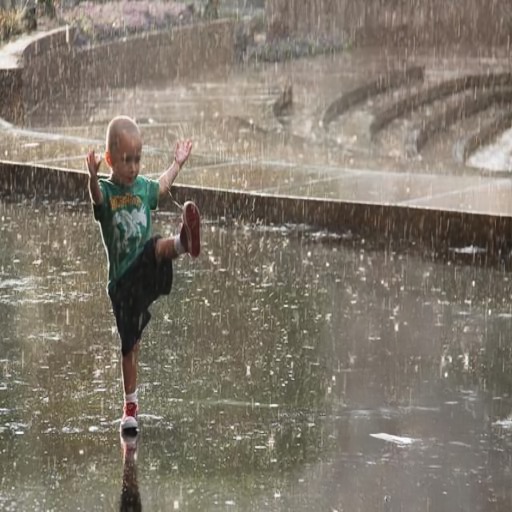}
    \hfill
    \includegraphics[width=1\linewidth]{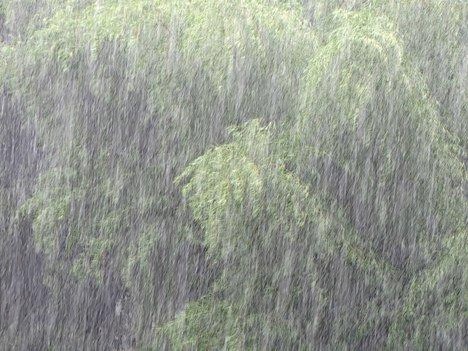}
    \end{minipage}
    }
  \hspace{-3.5mm}
    \subfloat[\cite{chen2023sparse}]{
  \begin{minipage}{0.140\linewidth}
    \centering
    \includegraphics[width=1\linewidth]{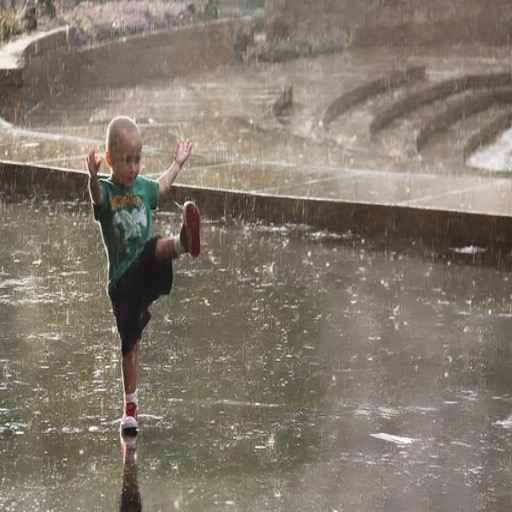}
    \hfill
    \includegraphics[width=1\linewidth]{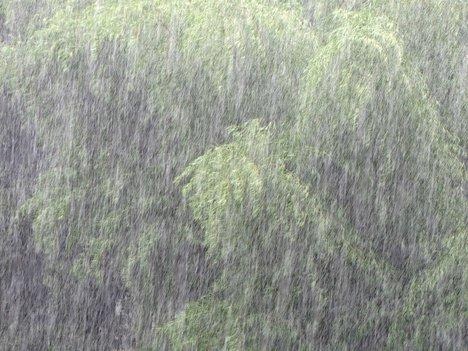}
    \end{minipage}
    }
  \hspace{-3.5mm}
    \subfloat[\cite{NeRD-Rain}]{
  \begin{minipage}{0.140\linewidth}
    \centering
    \includegraphics[width=1\linewidth]{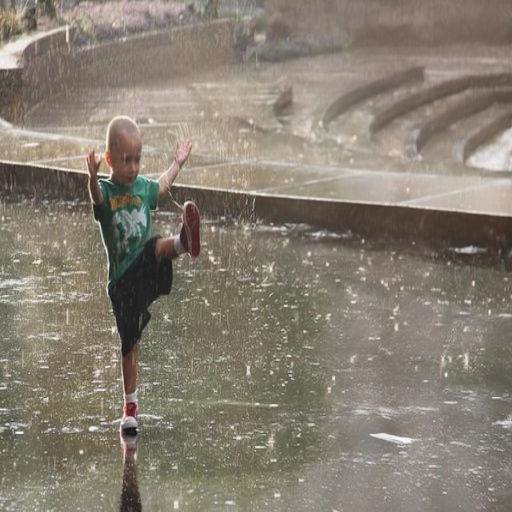}
    \hfill
    \includegraphics[width=1\linewidth]{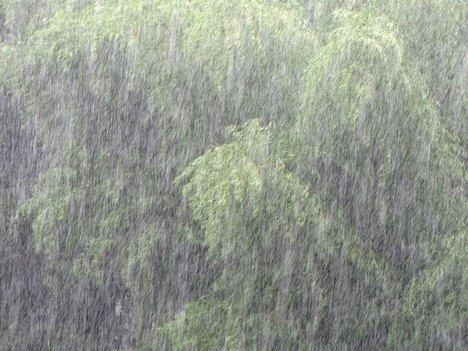}
    \end{minipage}
    }
  \hspace{-3.5mm}
    \subfloat[Ours]{
  \begin{minipage}{0.140\linewidth}
    \centering
    \includegraphics[width=1\linewidth]{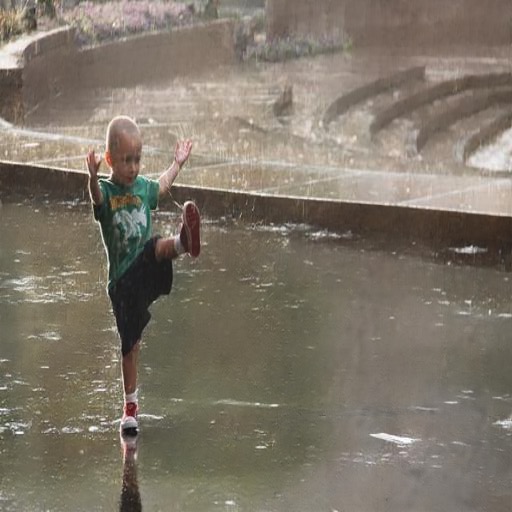}
    \hfill
    \includegraphics[width=1\linewidth]{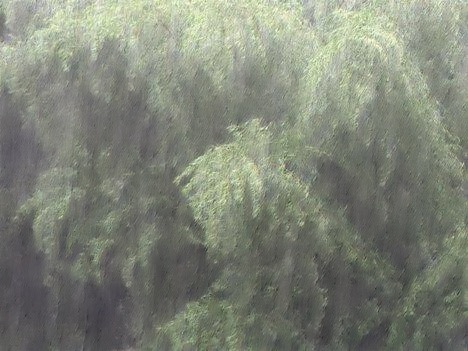}
    \end{minipage}
    }
  \caption{Visual comparisons of deraining on two samples from Internet-Data~\cite{wang2019spatial}. All model weights for real-world deraining are trained on SPA-Data~\cite{wang2019spatial}. The sample from (b) to (f) are Restormer~\cite{zamir2022restormer}, IDT~\cite{xiao2022image}, DRSformer~\cite{chen2023learning}, and UDR-S$^2$Former~\cite{chen2023sparse}, NeRD~\cite{NeRD-Rain}, respectively. Please zoom in for a better view.}
  \label{fig:real}
\end{figure*}

\subsection{Comparisons with the State-of-the-Arts}
\label{sec:results}

\paragraph{Quantitative Evaluation.}
In our study, we present a comprehensive comparative analysis on both synthetic and real rainy datasets, as summarized in Table~\ref{tab:result}. 
As indicated, our proposed method consistently outperforms all other deraining methods, achieving superior results in terms of PSNR. 
For instance, TransMamba surpasses the state-of-the-art approach UDR-S$^2$Fromer by an average of 0.32 dB.
The model size, and complexity, inference runtime and The non-reference image quality results on the unpaired dataset Internet-Data~\cite{wang2019spatial} are reported in the supplementary materials.

As the ground truths of real-world rainy images in Internet-Data~\cite{wang2019spatial} are missing, we conduct the qualitative comparison by computing non-reference image quality assessment metrics including NIQE~\cite{mittal2012making} and BRISQUE~\cite{mittal2012no} on 20 randomly sampled images.
The results are shown in Table~\ref{tab:niqe}.
As seen, our method achieves the best image qualities in terms of both non-reference metrics.

\paragraph{Qualitative Evaluation.}
Furthermore, we conduct a visual comparison on the six datasets, and the deraining results are depicted in Figures~\ref{fig:rain200lh}-\ref{fig:real}. Figure~\ref{fig:rain200lh} and Figure~\ref{fig:did&ddn} showcase the deraining results on synthetic datasets, while Figure~\ref{fig:spa} and Figure~\ref{fig:real} present the deraining outcomes on real-world datasets SPA-Data~\cite{wang2019spatial} and Internet-Data~\cite{wang2019spatial}, respectively. 
These visual results highlight the efficacy of our method in comprehensively eliminating rain degradation, including both small and large rain streaks, in both simulated and real-world rainy images.

\begin{table*}[htbp]
\footnotesize
\renewcommand\arraystretch{1.}
\tabcolsep=2.mm
  \centering
  \caption{The metrics comparison of real-world deraining involves non-reference image quality assessment on Internet-Data~\cite{wang2019spatial}.}
    \begin{tabular}{lrrrrrrrrrr}
    \toprule
          & \multicolumn{1}{l}{Rainy} & \multicolumn{1}{l}{SPDNet~\cite{yi2021Structure}} & \multicolumn{1}{l}{Uformer~\cite{wang2022uformer}} & \multicolumn{1}{l}{Restormer~\cite{zamir2022restormer}} & \multicolumn{1}{l}{IDT~\cite{xiao2022image}} & \multicolumn{1}{l}{DRSformer~\cite{chen2023learning}} & \multicolumn{1}{l}{UDR-S$^2$Former~\cite{chen2023sparse}} & \multicolumn{1}{l}{NeRD~\cite{NeRD-Rain}} & \multicolumn{1}{l}{Ours} \\
    \midrule
    NIQE\(\downarrow\)  & 3.66  & 3.88  & 4.21  & 3.62  & 3.71  & 4.03  & 3.52 & 3.49  & \textbf{3.31} \\
    BRISQUE\(\downarrow\) & 22.47 & 27.64 & 25.62 & 24.25 & 22.94 & 26.23 & 20.65 & 20.13 & \textbf{18.51} \\
    \bottomrule
    \end{tabular}%
  \label{tab:niqe}%
\end{table*}%

\subsection{Performance and Complexities}\label{sec::supple_complexities}
We also compare our method with the recent deraining models in terms of model size and computation complexities. 
Additionally, we present the deraining results in terms of PSNR on Rain200L~\cite{yang2017deep} for convenient reference to their performances.  
The values of complexities are shown in Table~\ref{tab:complexity}.
As shown, our method has comparable model size and computation overhead as the previous methods.
Note that the spectral-domain weights are defined in complex space and their complexities could be compressed by pruning them into real space.
We will tackle the issue of model efficiency based on the techniques in model compression like pruning and quantization. 
Generally, our method achieves a fair balance between the deraining performance and model complexity. 

\begin{table*}[htbp]
\footnotesize
\renewcommand\arraystretch{1}
\tabcolsep=2.5mm
  \centering
  \caption{The performance evaluations of deraining methods encompassing PSNR, model size, computational complexity, and runtime are conducted. PSNR values are assessed on Rain200L~\cite{yang2017deep}, and FLOPs results are computed on input images of dimensions 256$\times$256. The values of runtime are tested on a Nvidia Tesla V100 GPU.}
    \begin{tabular}{lrrrrrrrrr}
    \toprule
          &  \multicolumn{1}{l}{SPDNet~\cite{yi2021Structure}} & \multicolumn{1}{l}{Uformer~\cite{wang2022uformer}} & \multicolumn{1}{l}{Restormer~\cite{zamir2022restormer}} & \multicolumn{1}{l}{IDT~\cite{xiao2022image}} & \multicolumn{1}{l}{DRSformer~\cite{chen2023learning}} & \multicolumn{1}{l}{UDR-S$^2$Former~\cite{chen2023sparse}} & \multicolumn{1}{l}{NeRD~\cite{NeRD-Rain}} & \multicolumn{1}{l}{Ours} \\
    \midrule
    PSNR & 40.50 & 40.20 & 40.99 & 40.74 & 41.23 & 40.96 & 41.71 & 41.92 \\
    \#Parameter & 3.32M & 20.63M & 26.10M & 16.42M & 33.70M & 8.53M & 22.89M & 16.74M \\
    FLOPs & 96.29 & 43.86G & 140.99G & 61.90G & 242.9G & 21.58G & 156.3G & 45.67G \\
    Runtime & 0.24 & 0.19  & 0.28  & 0.28  & 0.59  & 0.12 & 0.42 & 0.13 \\
    \bottomrule
    \end{tabular}%
  \label{tab:complexity}%
\end{table*}%

\subsection{Ablation Studies}
\label{sec:ablation}

To assess the effectiveness of each ingredient within TransMamba, we conduct ablation studies on Rain200H~\cite{yang2017deep}. 
Specifically, we investigate the impact of the ingredients including
the self-attention choice,
the number of channels and bands in SBSA,
the feed-forward choice, and
the weight of spectral coherence loss.

\begin{figure}[!t]
\footnotesize
\renewcommand\arraystretch{1.}
\begin{minipage}{\linewidth}
\begin{minipage}[t]{0.494\linewidth}
\centering
\makeatletter\def\@captype{table}
\tabcolsep=1.5mm
    \caption{Ablation studies on the choice of the self-attention type.}\label{tab:abla_sa_type}%
    \begin{tabular}{lcc}
    \toprule
    Self-Attention & PSNR  & SSIM \\
    \midrule
    MDTA~\cite{zamir2022restormer}  & 32.37 & 0.9351 \\
    FSAS~\cite{kong2023efficient}  & 32.47 & 0.9364 \\
    TKSA~\cite{chen2023learning} & 32.54 & 0.9371 \\
    w/o SBR & 32.64 & 0.9383 \\
    SBSA  & \textbf{32.96} & \textbf{0.9409} \\
    \bottomrule
    \end{tabular}%
\end{minipage}
\hfill
\begin{minipage}[t]{0.494\linewidth}
\centering
\makeatletter\def\@captype{table}
\renewcommand\arraystretch{1}
\tabcolsep=2.5mm
    \caption{Ablation studies on the number of channels and bands.}\label{tab:abla_c_b}%
    \begin{tabular}{ccc}
    \toprule
    ($C_f,b$) & PSNR  & SSIM \\
    \midrule
    (72,1) & 32.74 & 0.9393 \\
    (36,2) & \textbf{32.96} & \textbf{0.9409} \\
    (18,4) & 32.78 & 0.9392 \\
    (9,8) & 32.33 & 0.9349 \\
    (4,18) & 31.72 & 0.9293\\
    \bottomrule
    \end{tabular}%
\end{minipage}
\end{minipage}
\vspace{-2mm}
\end{figure}

\begin{figure}[!t]
\footnotesize
\begin{minipage}{\linewidth}
\begin{minipage}[t]{0.494\linewidth}
\centering
\makeatletter\def\@captype{table}
\tabcolsep=1.5mm
    \caption{Ablation studies on feed-forward module}\label{tab:abla_ffn}%
    \begin{tabular}{lcc}
    \toprule
    Feed-Forward & PSNR  & SSIM \\
    \midrule
    FN~\cite{liang2021swinir}    & 32.47 & 0.9367 \\
    GDFN~\cite{zamir2022restormer}  & 32.58 & 0.9372 \\
    DFFN~\cite{kong2023efficient}  & 32.70 & 0.9384 \\
    MSFN~\cite{chen2023learning}  & 32.81 & 0.9391 \\
    SEFF  & \textbf{32.96} & \textbf{0.9409} \\
    \bottomrule
    \end{tabular}%
\end{minipage}
\hfill
\begin{minipage}[t]{0.494\linewidth}
\centering
\makeatletter\def\@captype{table}
\renewcommand\arraystretch{1}
\tabcolsep=1.5mm
  \caption{Ablation studies on state space module}\label{tab:abla_mamba}%
    \begin{tabular}{lcc}
    \toprule
    SSM & \multicolumn{1}{l}{PSNR} & \multicolumn{1}{l}{SSIM} \\
    \midrule
    VSSM~\cite{guo2024mambair} & 32.79 & 0.9388 \\
    For.+For. & 32.85 & 0.9402 \\
    Back.+Back. & 32.83 & 0.9399 \\
    Back.+For. & 32.94 & \textbf{0.941} \\
    For.+Back. & \textbf{32.96} & 0.9409 \\
    \bottomrule
    \end{tabular}%
\end{minipage}
\end{minipage}
\end{figure}

\input{sec/tab_tex/abla_loss}

\paragraph{SBSA.}
To substantiate the effectiveness of the proposed SBSA, we conduct a comparison with three baselines: 
the multi-Dconv head transposed attention (MDTA)~\cite{zamir2022restormer}, 
the frequency domain-based self-attention solver (FSAS)~\cite{kong2023efficient}, and 
the top-k sparse attention (TKSA)~\cite{chen2023learning}. 
Additionally, we explore an additional setting of SBSA by excluding the SBR step. 
The quantitative analysis results are presented in Table~\ref{tab:abla_sa_type}.
FSAS employs point-wise product in the frequency domain to replace conventional tensor multiplication during the generation of attention maps, resulting in the loss of global information that should have been captured by tensor multiplication. 
Although TKSA integrates rich local information across channels, it may lead to a loss in the extraction of long-range dependencies in the spectral domain. 
In contrast, our SBSA powered by SBR is capable of extracting long-range dependencies both in spectral and spatial domains, resulting in a PSNR improvement of 0.47 dB compared to TKSA.

\paragraph{Number of Channels and Bands.}
To evaluate the impact of the number of feature channels (\(C_f\)) and spectral bands (\(b\)) in SBSA, we conduct experiments with five different configurations, i.e., $(72,1)$, $(36,2)$, $(18,4)$, $(9,8)$, and $(4,18)$. 
Their product is constrained to be within $72$ due to limited memory. 
The results are presented in Table~\ref{tab:abla_c_b}. 
It is observed that, although decreasing the number of channels diminishes the learning capacity of the network, the configuration of $(18,4)$ still provides comparable performance to that of $(72,1)$, indicating the effectiveness of our SBSA. 
When the combination of (\(C_f\), \(b\)) reaches $(36,2)$, it leads to the best result among all the configurations.

\paragraph{SEFF.}
To evaluate the efficacy of the proposed SEFF, we conduct a comparative analysis with four alternatives:
(1) the vanilla feed-forward network (FN)~\cite{liang2021swinir},
(2) the gated-Dconv feed-forward network (GDFN)~\cite{zamir2022restormer},
(3) the discriminative frequency domain-based feed-forward network (DFFN)~\cite{kong2023efficient}, and
(4) the mixed-scale feed-forward network (MSFN)~\cite{chen2023learning}.
The quantitative results are shown in Table~\ref{tab:abla_ffn}.
While DFFN is capable of extracting frequency-based features, it falls short in leveraging multi-range spatial knowledge and enhancing frequency-specific information.
Despite MSFN integrating mixed-scale information, it may still overlook the exploitation of multi-range spatial knowledge and the feature exploitation in the spectral domain.
Through the incorporation of FFT and feature aggregation across different ranges in both spatial and spectral domains, our SEFF module further enhances performance, resulting in a PSNR gain of 0.14 dB over MSFN.

\paragraph{SSM.}
To evaluate the efficacy of the proposed CBSM, we conduct a comparative analysis with five settings:
(1) the Vision State-Space Module (VSSM)~\cite{guo2024mambair},
(2) the two cascaded forward-directional CBSM, 
(3) the two cascaded backward-directional CBSM, 
(4) the cascaded backward- and forward-directional CBSM, 
and 
(5) the cascaded forward- and backward-directional CBSM (ours).
As shown in Table~\ref{tab:abla_mamba}, the bi-directional settsings of (4) and (5) are the best.

\paragraph{Spectral Coherence Loss.}
Table~\ref{tab:abla_cohloss} illustrates the efficacy of the spectral coherence loss \(\mathcal{L}_{cor}\) and examines the impact of varying loss weight values. 
It is apparent that the incorporation of \(\mathcal{L}_{coh}\) consistently enhances performance, with the optimal result achieved when its loss weight \(\alpha\) attains a value of 5.



\begin{figure*}[tb]
  \centering
  \begin{minipage}{0.465\linewidth}
  \subfloat[Input]{
    \includegraphics[trim=35mm 81mm 35mm 52mm,clip,width=1\linewidth]{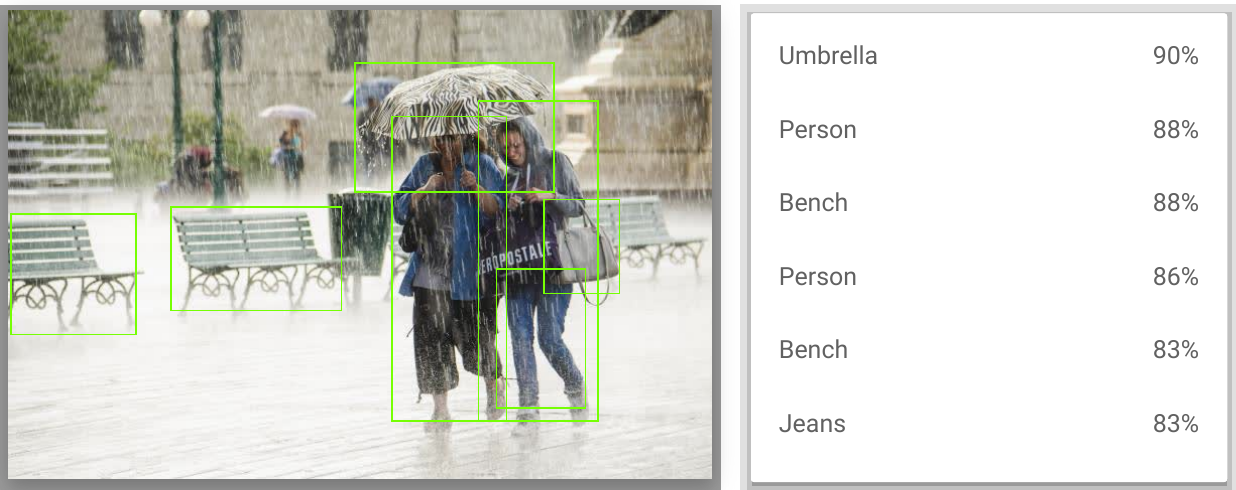}
    }
  \end{minipage}
  \hspace{-2mm}
  \begin{minipage}{0.465\linewidth}
  \subfloat[Derained by ours]{
    \includegraphics[trim=35mm 81mm 35mm 52mm,clip,width=1\linewidth]{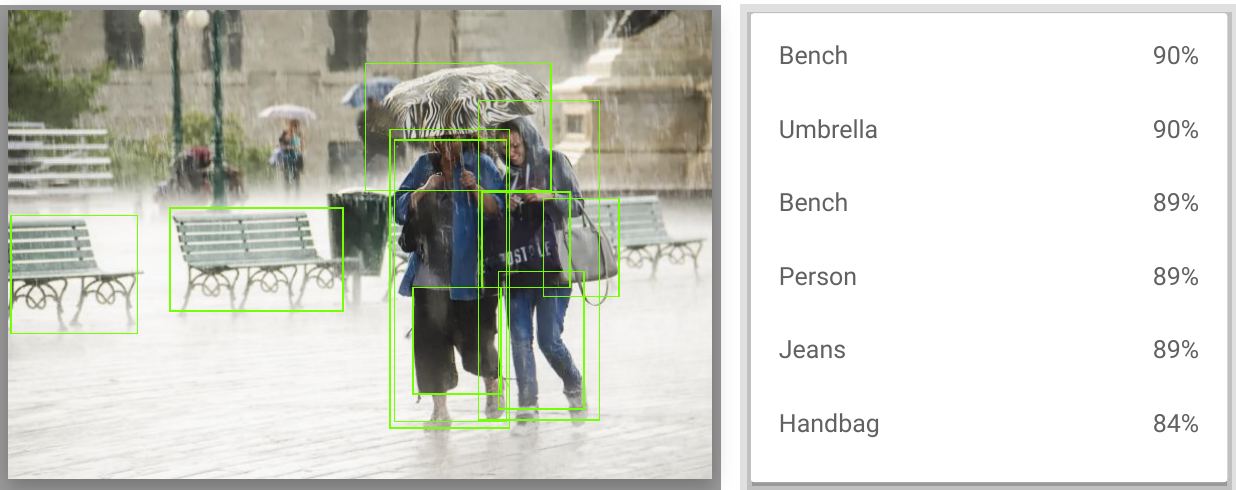}
    }
  \end{minipage}

  \begin{minipage}{0.465\linewidth}
  \subfloat[Input]{
    \includegraphics[trim=35mm 76mm 35mm 52mm,clip,width=1\linewidth]{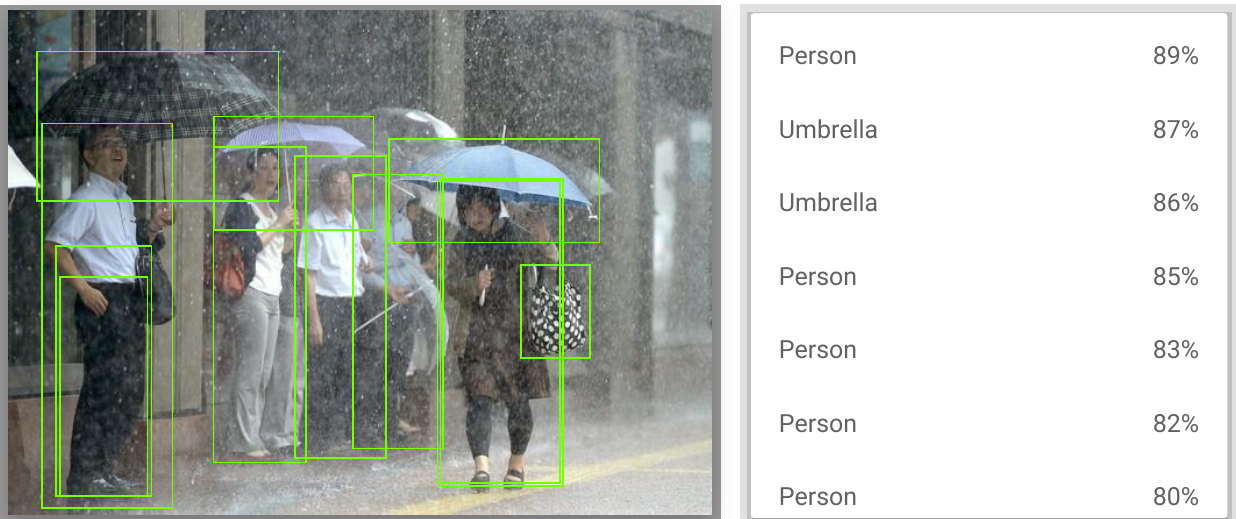}
    }
    \label{fig:real_detect-a}
  \end{minipage}
  \hspace{-2mm}
  \begin{minipage}{0.465\linewidth}
  \subfloat[Derained by ours]{
    \includegraphics[trim=35mm 76mm 35mm 52mm,clip,width=1\linewidth]{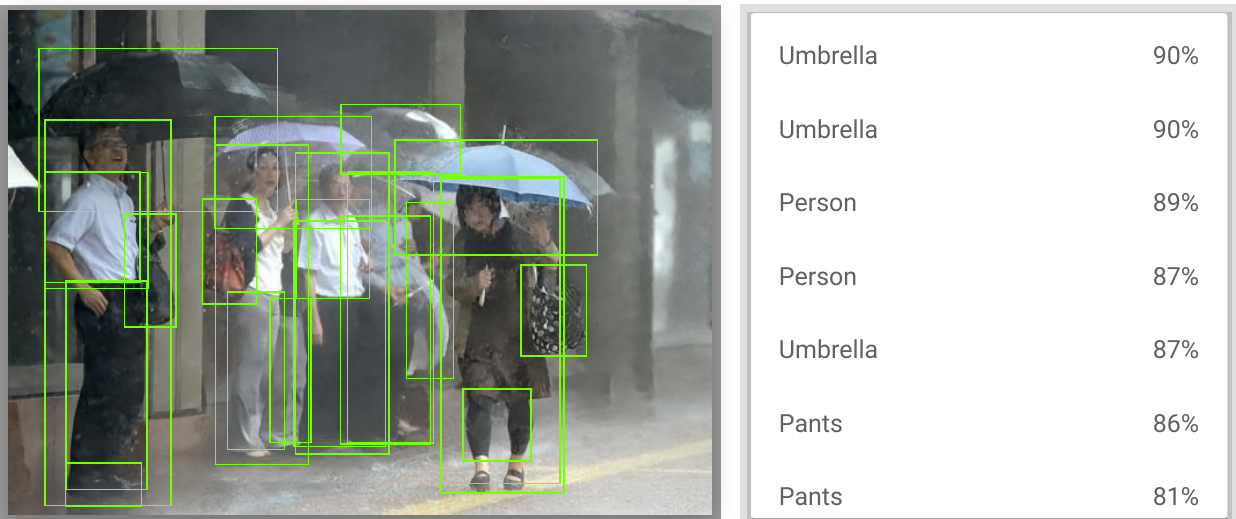}
    }
    \label{fig:real_detect-b}
  \end{minipage}
  \caption{Deraining result on two real-world rainy images from Internet-Data~\cite{wang2019spatial} and their detection results by Google API.}
  \label{fig:real_detect}
\end{figure*}
\begin{table}[t]
\tabcolsep=.7mm
  \centering
  \caption{Deraining and subsequent detection results by a pretrained YOLOv3 on \textbf{COCO350/BDD350}.}
    \begin{tabular}{lcccc}
    \toprule
        & Rain  & DRSformer & UDR-S2Former & Ours \\
    \midrule
    Precision (\%)  & 24.13/37.56  & 33.41/41.29 & 33.58/41.65 & \textbf{34.04/41.77} \\
    Recall (\%)  & 29.82/41.93  & 39.97/50.74 & 40.23/51.04 & \textbf{40.99/52.17} \\
    IoU (\%)  & 56.51/60.44  & 62.03/62.14 & 62.32/62.47 & \textbf{62.82/62.97} \\
    \bottomrule
    \end{tabular}%
  \label{tab:bdd_coco}%
\end{table}%

\begin{figure*}[tb]
  
  \begin{minipage}{0.197\linewidth}
    \centering
    \begin{annotationimage}{width=1\linewidth}{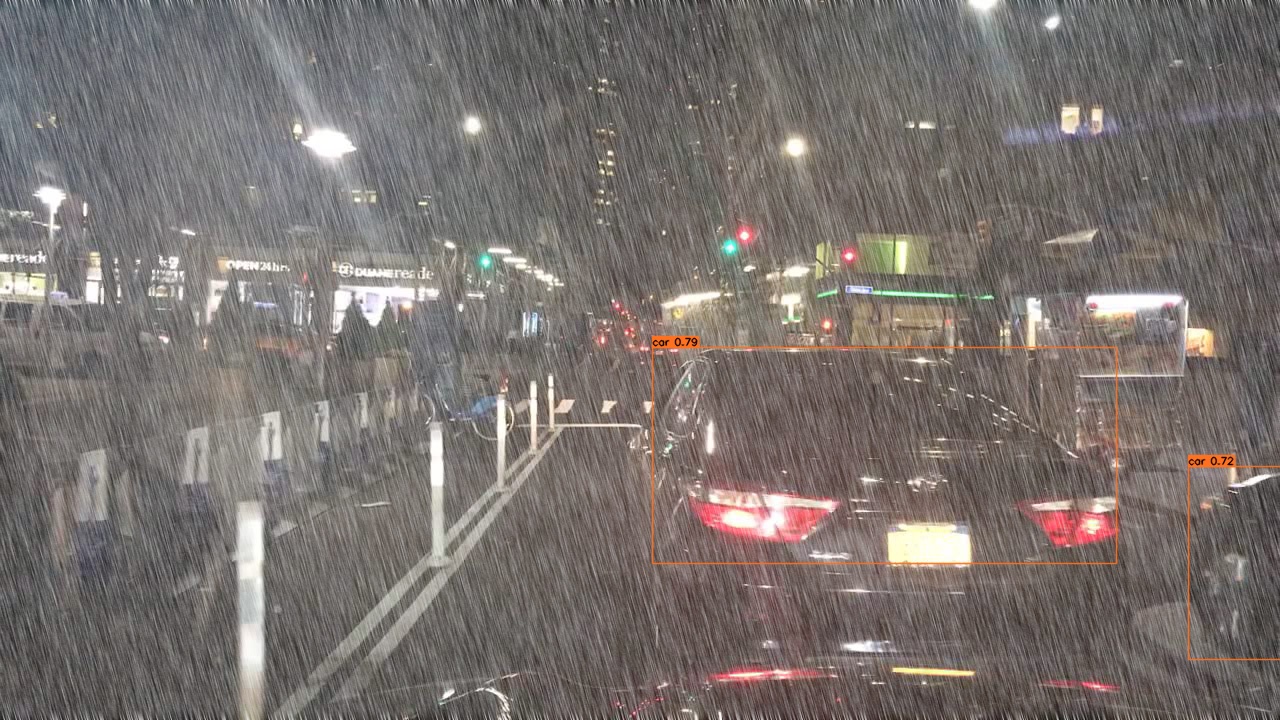}
    \end{annotationimage}
    \end{minipage}
  \hspace{-1.65mm}
  \begin{minipage}{0.197\linewidth}
    \centering
    \begin{annotationimage}{width=1\linewidth}{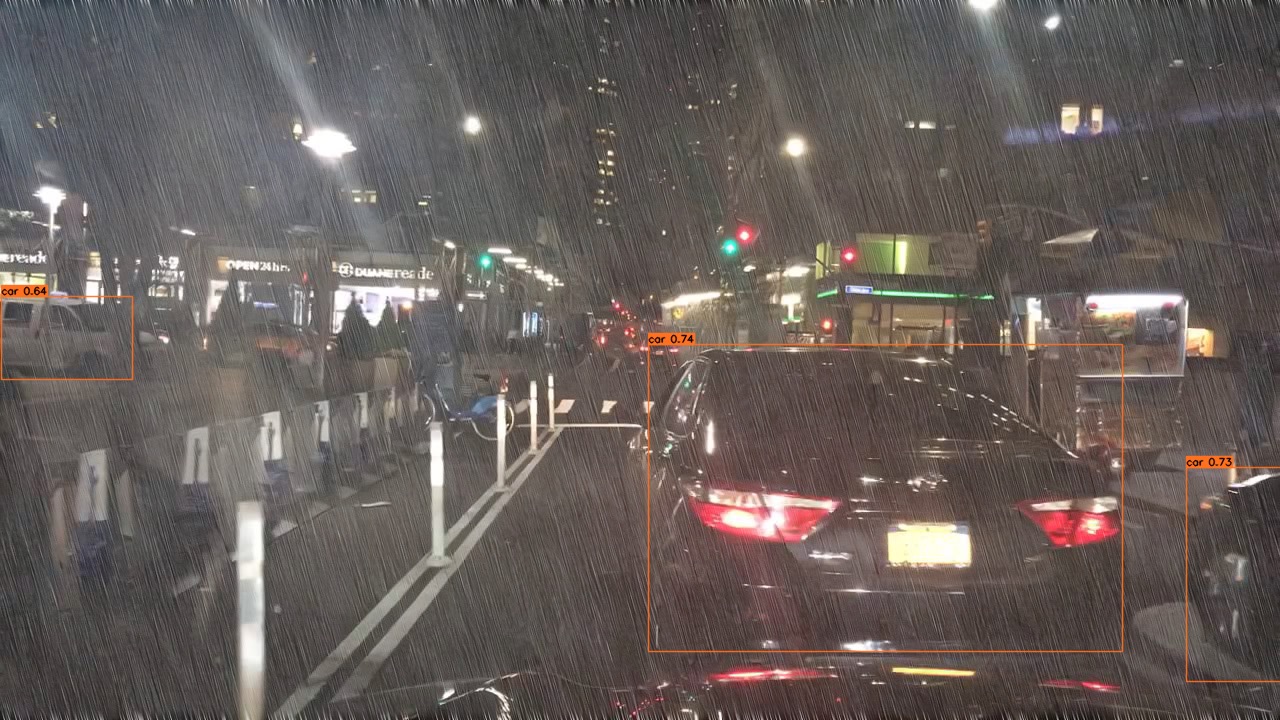}
    \end{annotationimage}
    \end{minipage}
  \hspace{-1.65mm}
  \begin{minipage}{0.197\linewidth}
    \centering
    \begin{annotationimage}{width=1\linewidth}{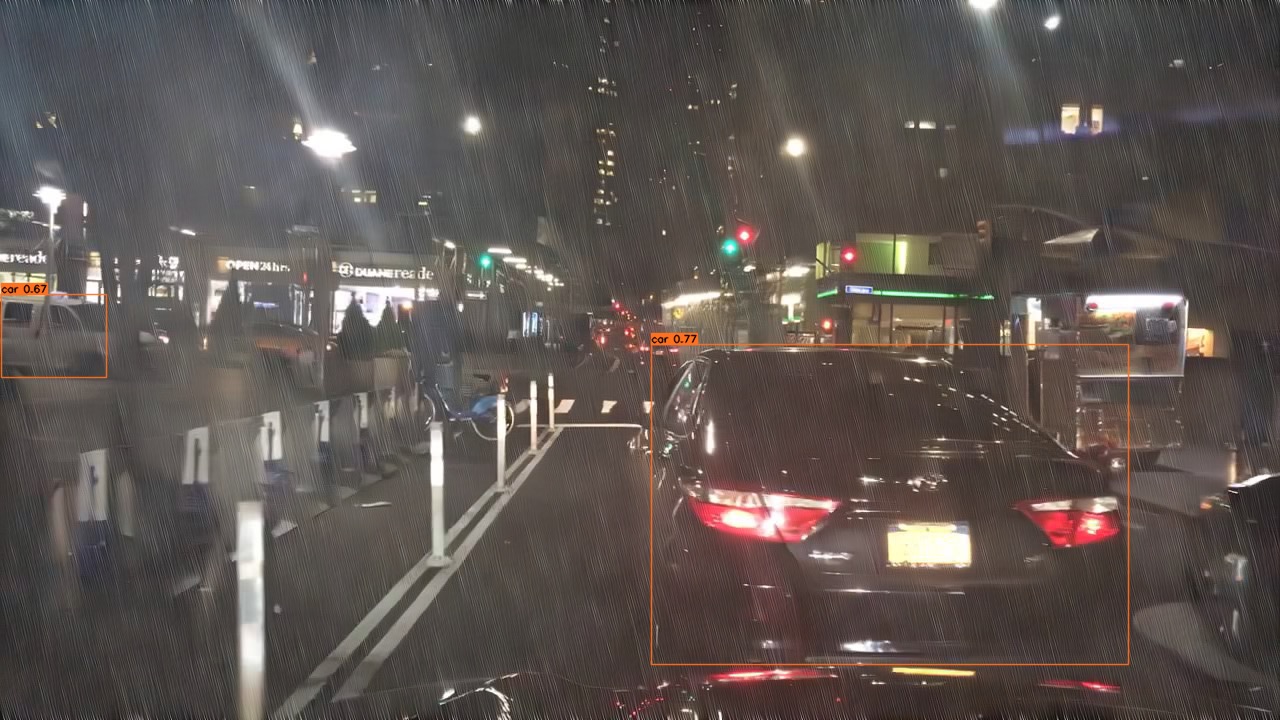}
    \end{annotationimage}
    \end{minipage}
  \hspace{-1.65mm}
  \begin{minipage}{0.197\linewidth}
    \centering
    \begin{annotationimage}{width=1\linewidth}{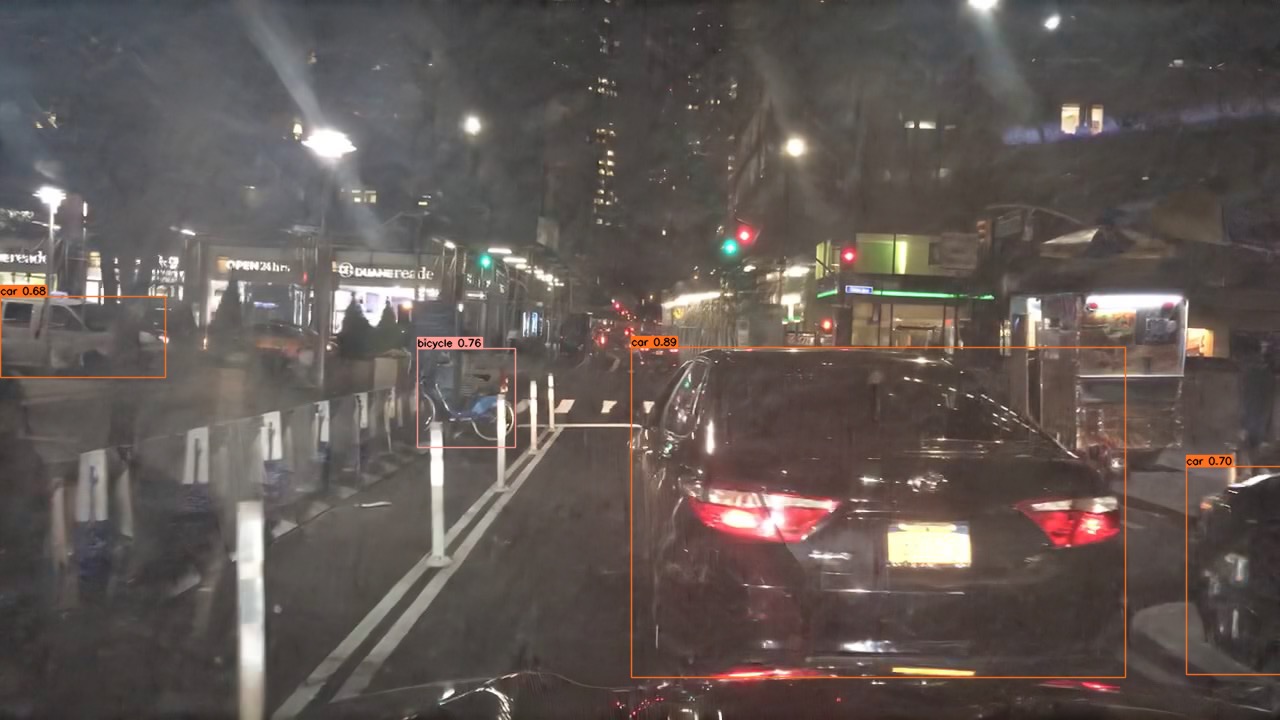}
    \end{annotationimage}
    \end{minipage}
  \hspace{-1.65mm}
  \begin{minipage}{0.197\linewidth}
    \centering
    \begin{annotationimage}{width=1\linewidth}{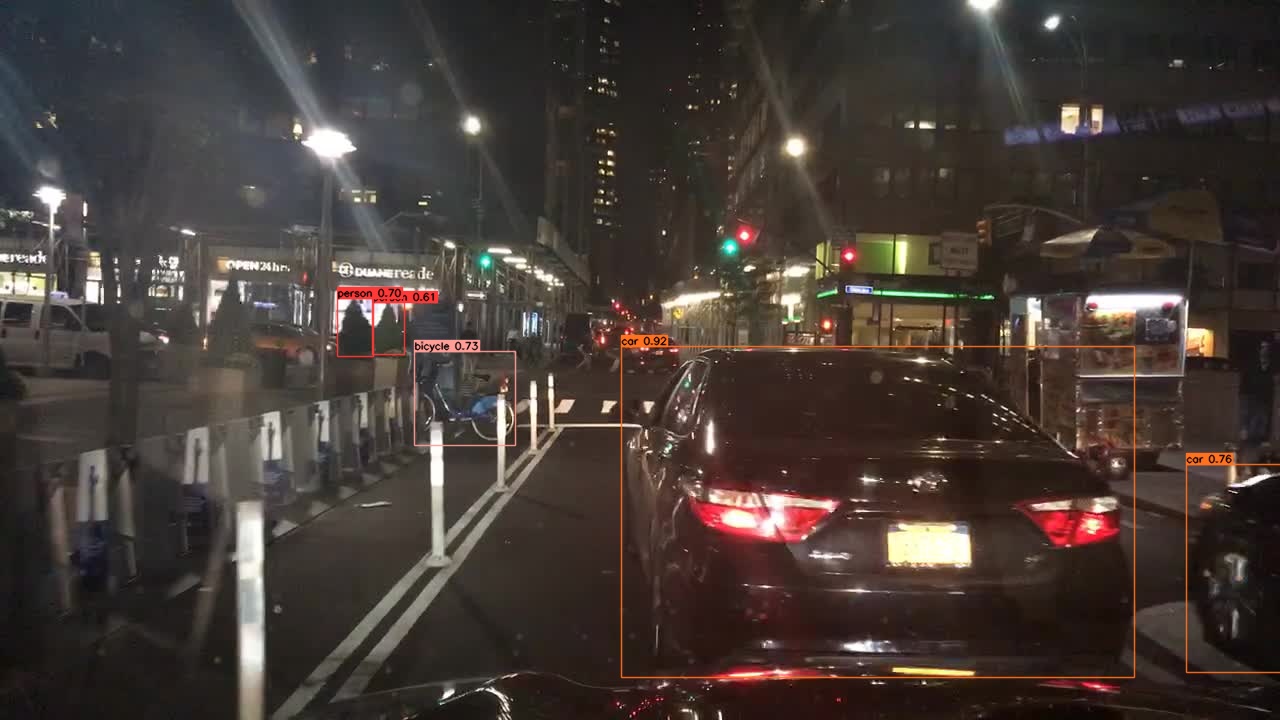}
    \end{annotationimage}
    \end{minipage}
    
  \begin{minipage}{0.197\linewidth}
    \centering
    \begin{annotationimage}{width=1\linewidth}{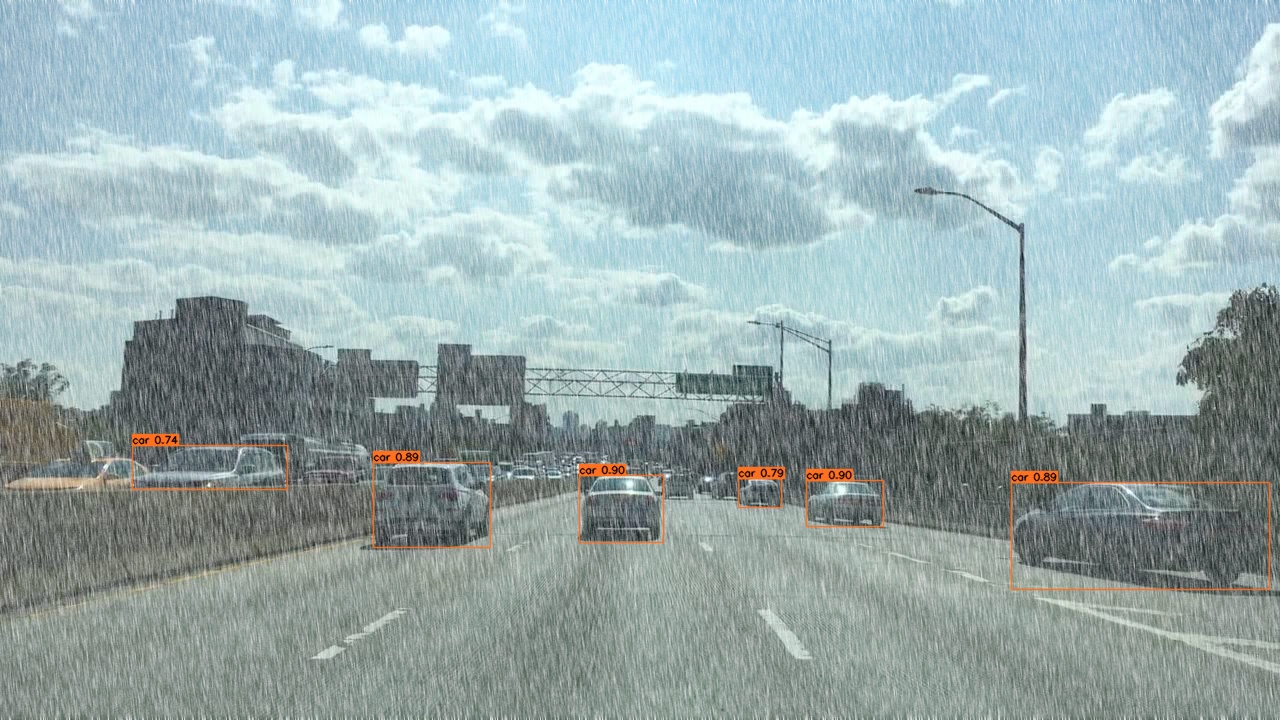}
    \node[fill opacity=1,text=white] at (0.5,0.11) {\scriptsize Input};
    \end{annotationimage}
    \end{minipage}
  \hspace{-1.65mm}
  \begin{minipage}{0.197\linewidth}
    \centering
    \begin{annotationimage}{width=1\linewidth}{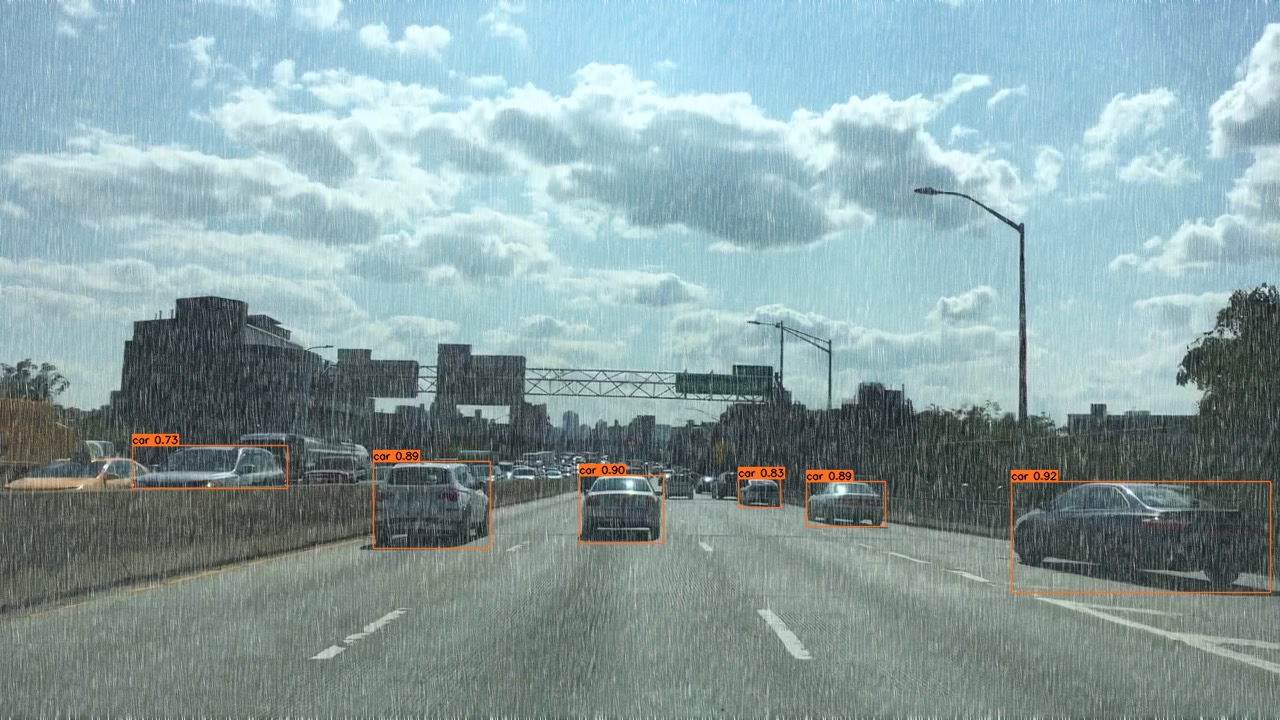}
    \node[fill opacity=1,text=white] at (0.5,0.11) {\scriptsize DRSformer};
    \end{annotationimage}
    \end{minipage}
  \hspace{-1.65mm}
  \begin{minipage}{0.197\linewidth}
    \centering
    \begin{annotationimage}{width=1\linewidth}{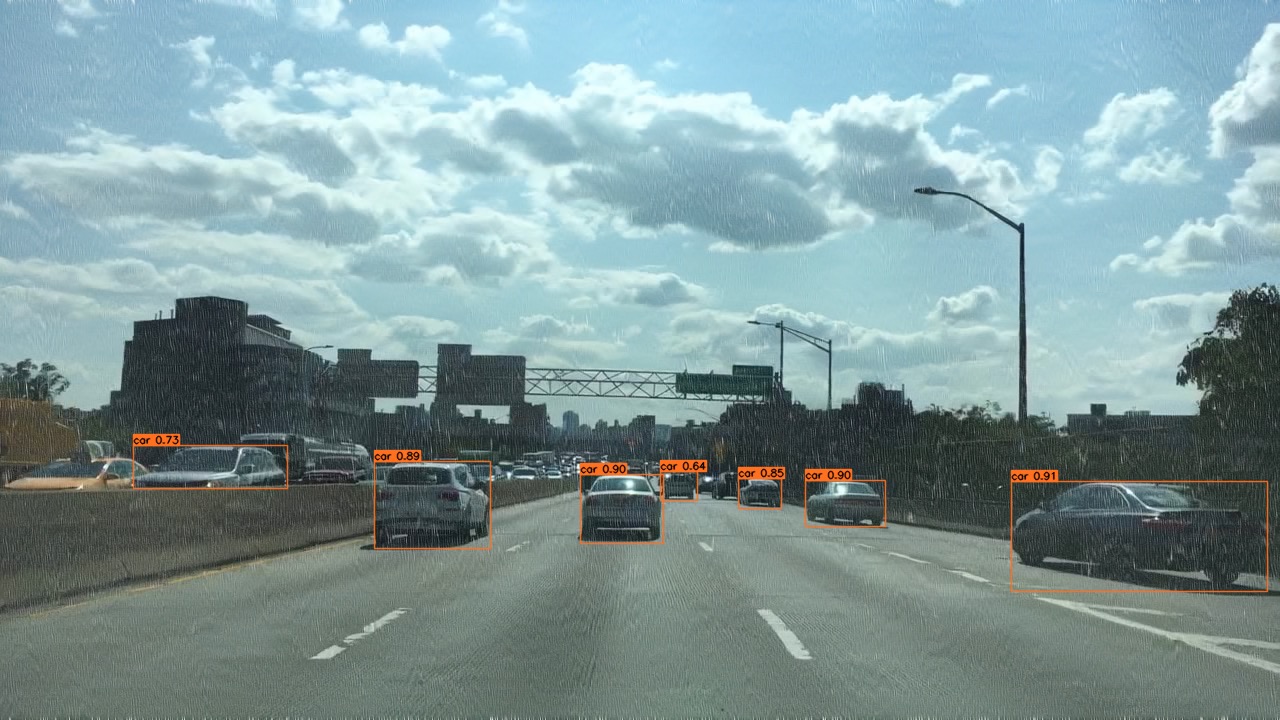}
    \node[fill opacity=1,text=white] at (0.5,0.11) {\scriptsize UDR-S$^2$Former};
    \end{annotationimage}
    \end{minipage}
  \hspace{-1.65mm}
  \begin{minipage}{0.197\linewidth}
    \centering
    \begin{annotationimage}{width=1\linewidth}{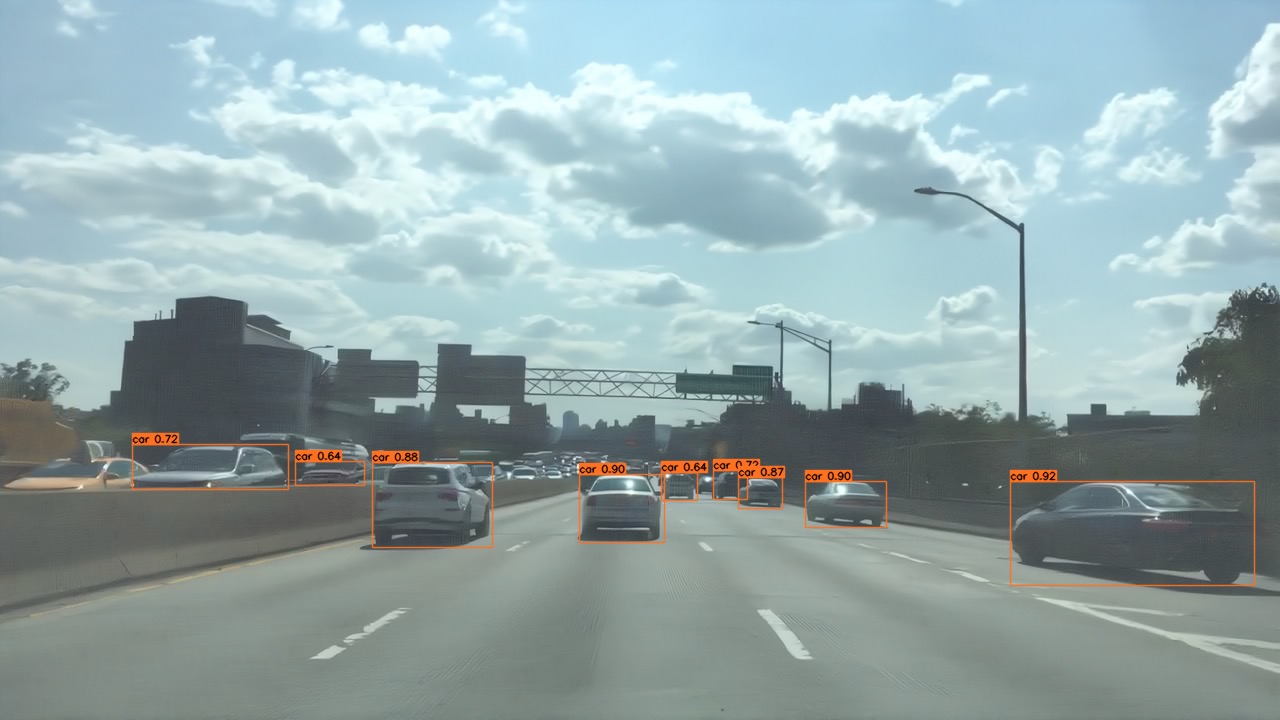}
    \node[fill opacity=1,text=white] at (0.5,0.11) {\scriptsize Ours};
    \end{annotationimage}
    \end{minipage}
  \hspace{-1.65mm}
  \begin{minipage}{0.197\linewidth}
    \centering
    \begin{annotationimage}{width=1\linewidth}{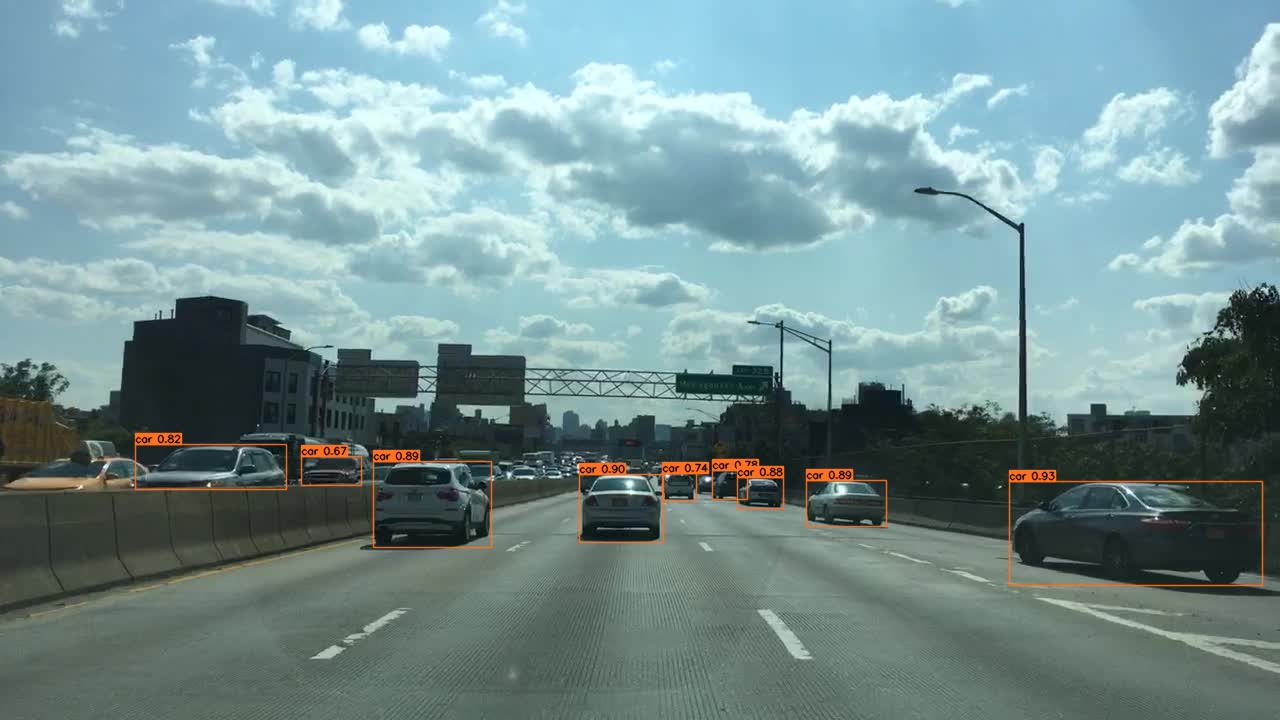}
    \node[fill opacity=1,text=white] at (0.5,0.11) {\scriptsize Clean};
    \end{annotationimage}
    \end{minipage}
  \caption{\footnotesize Comparison of deraining and detection by YOLOv3. Only ours removes all rain streaks and helps detect ``bicycle'' and 2 more ``cars''.}
  \label{fig:bdd_coco}
\end{figure*}

\subsection{Real-World Application}

To further show the practical application of TransMamba in real-world deraining and its capacity to enhance the downstream detection task, we present two examples in Figure~\ref{fig:real_detect}. 
As illustrated, all rain streaks are removed from the scene, thereby aiding the detector in accurately identifying previously obscured objects such as shoes and handbags.

\paragraph{Downstream Task.}
To more extensively assess the impact of image deraining on downstream vision application of object detection, we utilize the popular object detection model YOLOv3~\cite{redmon2018yolov3} to evaluate the derained results on COCO350 and BDD350 datasets~\cite{jiang2020multi,jiang2023dawn}. 
Table~\ref{tab:bdd_coco} illustrates that our TransMamba yields the best object detection performance among deraining methods. 
Visual comparisons of two instances in Figure~\ref{fig:bdd_coco} demonstrate that TransMamba derained images exhibit the best results in image quality and detection accuracy. 

\subsection{Limitation and Future Work}
Due to the powerful ability of removing rain accumulation, our TransMamba may make restored background over-smooth. We will leverage diffusion model to alleviate the issue. Besides, we are developing algorithms of model compression and acceleration to enable the real-time usage of rain streak removal models.

\section{Conclusion}

In this work, we propose a spectral-domain banding Transformer to efficiently remove rain streaks in single images. 
We leverage the inherent decomposed information of spectral-domain tokens for capturing global dependencies and thus propose self-attention in spectral space. 
Observing that spectral bands intrinsically separate backgrounds of detailed textures and rain streaks of repeated patterns, we propose a spectral banding self-attention to allocate varying attention to different bands. 
Additionally, we present a spectral enhanced feed-forward module for better extracting frequency-specific information. 
To better reconstruct innate relations within clean image signals, we also develop a spectral coherence loss.
Extensive experiments on both synthetic benchmarks and real-world rainy images show the superiority of our method.

\bibliographystyle{IEEEtran}
\bibliography{main}

\vfill

\end{document}